\documentclass{article}

\PassOptionsToPackage{numbers,sort&compress}{natbib}
\usepackage[main, preprint]{neurips_2026}

\usepackage[utf8]{inputenc}
\usepackage[T1]{fontenc}
\usepackage{microtype}
\usepackage{graphicx}
\usepackage[hypertexnames=false,bookmarks=false,breaklinks=true]{hyperref}
\makeatletter
\let\hyper@natlinkstart\@gobble
\let\hyper@natlinkend\@empty
\def\hyper@natlinkbreak#1#2{#1}
\makeatother
\usepackage{url}
\usepackage{amsmath}
\usepackage{amssymb}
\usepackage{amsfonts}
\usepackage{bm}
\usepackage{amsthm}
\usepackage{nicefrac}
\usepackage{xcolor}
\usepackage{booktabs}
\usepackage{multirow}
\usepackage{algorithm}
\usepackage{algorithmic}
\makeatletter
\renewcommand{\@noticestring}{%
  Preprint. Correspondence to Yao Shu \texttt{<yaoshu@hkust-gz.edu.cn>}.%
}
\makeatother
\graphicspath{{workspace/figs/}}
\newcommand{\R}{\mathbb{R}}
\newcommand{\E}{\mathbb{E}}
\newcommand{\sgn}{\mathrm{sgn}}

\newcommand{\defeq}{\triangleq}
\newcommand{\hgrad}{\hat{\nabla}}

\newcommand{\vx}{\bm{x}}
\newcommand{\vy}{\bm{y}}
\newcommand{\vz}{\bm{z}}
\newcommand{\vs}{\bm{s}}
\newcommand{\va}{\bm{a}}
\newcommand{\vg}{\bm{g}}
\newcommand{\vu}{\bm{u}}

\newcommand{\vr}{\bm{r}}
\newcommand{\ve}{\bm{e}}
\newcommand{\vp}{\bm{p}}
\newcommand{\vb}{\bm{b}}
\newcommand{\mI}{\bm{I}}
\newcommand{\mM}{\bm{M}}

\newtheoremstyle{paperplain}
  {3pt plus 1pt minus 1pt}
  {3pt plus 1pt minus 1pt}
  {\fontfamily{LibertinusSerif-TLF}\selectfont\itshape}
  {}
  {\fontfamily{LibertinusSerif-TLF}\selectfont\bfseries}
  {.}
  {0.5em}
  {}
\theoremstyle{paperplain}
\newtheorem{theorem}{Theorem}

\newtheorem{lemma}{Lemma}
\theoremstyle{definition}

\newtheorem{assumption}{Assumption}
\theoremstyle{remark}

\title{Compander-Aligned Query Geometry for Quantized Zeroth-Order Optimization}

\author{Yao Shu \quad Zilin Zhu}

\begin{document}
\raggedbottom
\maketitle

\begin{abstract}
Low-bit forward evaluation is an attractive route to memory-efficient zeroth-order (ZO) adaptation: the optimizer needs only scalar losses, and the model can be queried near deployment precision. The obstacle is that a quantized ZO query is not a continuous finite difference followed by harmless storage rounding. The query chooses endpoints, the low-precision engine rounds them, and the loss difference is measured along the rounded chord. For nonuniform companding quantizers, this makes the codebook insufficient to predict ZO behavior: a fixed weight-space radius can collapse in dense cells, over-span sparse cells, or assign a rounded chord to an unrounded update direction. We identify the missing object as query geometry and model scalar nonuniform quantization as \(Q=\phi^{-1}\circ U\circ\phi\). CAQ-ZO (Compander-Aligned Queries for Zeroth-Order Optimization) forms one-grid-step Rademacher stencils \(\vz\pm\Delta\vr\) in \(z=\phi(x)\), maps endpoints back through \(\phi^{-1}\), and updates in \(\vz\). Our theory proves the grid-span mismatch, decomposes endpoint-rounding estimator residuals, and gives stationarity bounds in which generic off-grid queries retain a \(\Delta^2/\mu^2\) residual channel while CAQ-ZO makes the query-time residual exactly zero. Synthetic experiments isolate this channel, and matched NF4 Qwen/Llama fine-tuning shows that CAQ-ZO improves the trained NF4 baseline under the same quantizer and evaluation budget.
\end{abstract}

\section{Introduction}

Zeroth-order optimization builds updates from function-value queries \citep{spall1992spsa,duchi2015zero,nesterov2017random,ghadimi2013zeroth}; in large-model training, MeZO uses this forward-only structure to operate near inference memory cost \citep{malladi2023mezo}. Quantization offers the complementary memory lever: if the model is stored and queried in low precision, adaptation can in principle inherit the storage footprint of deployment. This combination is especially tempting for large language models, where LoRA/QLoRA-style adaptation \citep{hu2021lora,dettmers2023qlora} and low-bit training refinements such as LoftQ \citep{li2024loftq} and LR-QAT \citep{liu2024lrqat} have made low-bit adaptation a standard practical target. The natural hope is therefore simple: run forward-only ZO through a low-precision model and keep the memory profile close to deployment. This hope hides a measurement problem. In continuous ZO, the estimator asks for losses at \(\vx\pm\mu\vu\). Under low-precision forward passes, the engine evaluates the rounded endpoints \(Q(\vx\pm\mu\vu)\) before the loss difference is formed. The central question is therefore not only which codebook reconstructs weights well, but which finite-difference endpoints make the quantized measurement informative.

This question matters most for nonuniform quantization. Nonuniform scalar quantizers include Lloyd-Max codebooks \citep{max1960quantizing,lloyd1982least}, companding quantizers \citep{gray1998quantization}, and NF4-like quantile codebooks for low-bit neural weights \citep{dettmers2023qlora}; all allocate more levels where values are statistically dense. For reconstruction or inference, this allocation can be exactly the right decision. For ZO, however, a fixed weight-space smoothing radius is interpreted through the local cell lengths of the quantizer. The same radius can collapse inside dense cells, act as a local secant near a matched cell scale, or cross many cells and become a nonlocal secant. In vector queries, the numerator is measured along the rounded chord \(Q(\vx+\mu\vu)-Q(\vx-\mu\vu)\), while the response is still assigned to \(\vu\). Thus the failure is not merely larger static quantization error; it is query-time quantization-perturbation coupling, a measurement channel that can bias the estimator and distort its descent direction.

We answer this bottleneck by making \emph{query geometry} the design object. By query geometry, we mean the finite-difference stencil relative to the quantizer cells: where the two endpoints are formed, how far and in which directions they move, and whether they lie on the quantization grid. We write scalar nonuniform quantization as a companding quantizer \(Q=\phi^{-1}\circ U\circ\phi\), where \(U\) is a uniform grid in the coordinate \(z=\phi(x)\). This representation separates the codebook from the coordinate in which the query is formed. Weight-space perturbations inherit position-dependent cell lengths. CAQ-ZO instead forms a one-grid-step Rademacher stencil \(\vz\pm\Delta\vr\) in the compander coordinate, maps the endpoints back through \(\phi^{-1}\), and updates \(\vz\) directly. Because the queried \(z\)-endpoints already lie on the uniform grid, the low-precision quantizer is the identity at query time and the measured population response is a finite difference of \(F\circ\phi^{-1}\). CAQ-ZO is therefore not a new codebook, straight-through estimator, or optimizer wrapper; it is the compander-aligned query geometry for the same low-precision forward oracle.

\textbf{Contributions.} This paper makes four contributions. First, it formulates low-precision-forward-pass ZO with the codebook, query coordinate, direction distribution, smoothing radius, and endpoint grid alignment kept as separate controls, turning quantized ZO from a codebook comparison into a measurement-geometry problem. Second, it develops the theory of this problem: a normalized compander-grid span theorem shows why no fixed weight-space radius can globally calibrate nonuniform cells; the rounded-chord and endpoint-residual analysis shows how this mismatch enters estimator error; and the stationarity comparison makes the resulting residual floor explicit for weight-space queries while proving that CAQ-ZO removes the query-time residual exactly. Third, it gives the CAQ-ZO algorithm, which uses on-grid Rademacher finite differences in \(z\), direct \(z\)-updates, and update-time projection to preserve on-grid future queries. Fourth, controlled synthetic experiments isolate the endpoint-rounding channel under fixed companders, and matched NF4 Qwen/Llama fine-tuning tests the target low-precision-forward-pass setting under the same quantizer and evaluation budget. The claims are scoped to scalar monotone companders and this controlled evidence, not to universal large-model transfer claims.

\section{Preliminaries}
\label{sec:preliminaries}

\subsection{Problem Setup}

We study stochastic zeroth-order optimization through low-precision forward evaluation, following the forward-only optimization view used in memory-efficient ZO \citep{malladi2023mezo} and recent quantized-ZO formulations \citep{feng2024stepping,zhou2025quzo}. The optimizer chooses a weight vector \(\vx\in\R^d\), and the optimization target is the smooth population loss
\begin{equation}
\min_{\vx\in\R^d} F(\vx),
\qquad
F(\vx)\defeq\E_{\xi}\!\left[f(\vx;\xi)\right].
\end{equation}
Here \(f(\cdot;\xi)\) is the stochastic forward loss and \(\xi\) denotes data, minibatch, task, or simulation randomness. The low-precision engine, however, does not expose \(f(\vy;\xi)\) at an arbitrary submitted endpoint \(\vy\); it evaluates the quantized state and returns \(f(Q(\vy);\xi)\). Thus \(F\circ Q\) is the measured population response of a weight-space query, not the final stationarity target. The optimizer has no access to sample or population gradients, nor to a straight-through gradient through \(Q\); it only receives scalar losses from this quantized measurement channel. This is the forward-only setting of 4- or 8-bit adaptation without backpropagation or high-precision optimizer state. Since every update is inferred from loss differences after quantized evaluation, \(Q\) is part of the finite-difference measurement rather than a storage detail added after optimization. The first query-geometry choice is therefore how to form the two endpoints before they enter this low-precision measurement.

\subsection{Weight-Space Quantized Finite Differences}

The weight-space baseline keeps the standard SPSA and two-point finite-difference stencil \citep{spall1992spsa,ghadimi2013zeroth} used in randomized-smoothing ZO \citep{duchi2015zero,nesterov2017random}. At weight \(\vx\), it samples directions, forms the weight-space stencil \(\{\vx-\mu\vu_k,\vx+\mu\vu_k\}\), and evaluates those endpoints after quantization. With smoothing radius \(\mu>0\), directions \(\vu_1,\ldots,\vu_K\), and one stochastic sample \(\xi\) shared by all \(2K\) evaluations, the measured estimator is
\begin{equation}
\label{eq:weight-space-qzo-estimator}
\hgrad(F\circ Q)(\vx)
\defeq
\frac1K\sum_{k=1}^K
\frac{
f(Q(\vx+\mu\vu_k);\xi)-f(Q(\vx-\mu\vu_k);\xi)
}{2\mu}\,\vu_k .
\end{equation}
The \(F\) in \(\hgrad(F\circ Q)\) names the smooth population loss being measured through quantized endpoints; the endpoint values use \(f\) because they are the observable function-value calls. The random sample is carried by the endpoint losses \(f(\cdot;\xi)\), not by an extra estimator subscript. The absence of a direction argument denotes the displayed average, and the \(k\)th sampled summand is written as \(\hgrad(F\circ Q)(\vx;\vu_k)\). Sharing \(\xi\) keeps the loss difference from being dominated by avoidable minibatch mismatch. The case \(K=1\) gives the single-direction estimate; Gaussian, sign, and Rademacher variants change the direction distribution while preserving the same weight-space stencil. The defining feature of Eq.~\eqref{eq:weight-space-qzo-estimator} is the order of operations: the perturbation is generated in weight space, and \(Q\) is applied afterward to both endpoints. For a uniform grid this order has one global resolution scale. For a nonuniform companding quantizer, the same \(\mu\) can be small relative to some local cells and large relative to others, so the estimator is determined by the query geometry as well as the direction distribution.

\subsection{Companding Quantizers and Query Geometry}

A standard model for a broad class of nonuniform scalar quantizers is companding \citep{jayant1984digital,gray1998quantization}: apply a monotone coordinate map, quantize uniformly in that coordinate, and map back. For the class studied here, this gives
\begin{equation}
Q(\vx)\defeq\phi^{-1}\!\left(U(\phi(\vx))\right),\qquad \vz=\phi(\vx).
\end{equation}
Here \(\phi\) is applied coordinatewise, \(\vz\) is the compander coordinate, and \(U\) is a uniform quantizer with spacing \(\Delta\). Uniform quantization is the special case \(\phi=\mathrm{id}\). Classical \(\mu\)-law and A-law quantizers are exact companders \citep{jayant1984digital,itu1988g711}; table codebooks such as NF4 can be connected to this notation through monotone interpolation \citep{dettmers2023qlora}, as discussed in Appendix~\ref{app:compander-families}, with non-scalar limitations in Appendix~\ref{app:beyond-companders}. Away from clipping, round-to-nearest quantization in the \(z\)-coordinate has residual at most \(\Delta/2\). The companding representation separates the oracle response, the smooth reference loss, and the coordinate in which endpoints are formed. The low-precision oracle returns \(f(Q(\vx);\xi)\), whose expectation is \(F(Q(\vx))\). The smooth reference loss \(F(\vx)\) is not queried directly; it exposes the measurement error caused by endpoint rounding. In the compander coordinate, an unrounded endpoint \(\vs\) has reference population value \(F(\phi^{-1}(\vs))\), while the same low-precision engine returns \(f(\phi^{-1}(U(\vs));\xi)\), whose expectation is \(F(\phi^{-1}(U(\vs)))=F(Q(\phi^{-1}(\vs)))\). Thus a compander-coordinate query changes neither the loss nor the codebook; it changes where the two finite-difference endpoints are formed before the fixed quantized measurement is applied.

This separation fixes the comparison axes. Weight-space ZO forms \(\vx\pm\mu\vu\) and may choose Gaussian, sign, or Rademacher directions; its endpoints are generally rounded after the perturbation. CAQ-ZO forms endpoints in the \(z\)-coordinate and uses a Rademacher one-grid-step stencil because exact on-grid endpoints require each coordinate to move by a signed grid increment. The controlled variables are therefore the codebook, query coordinate, direction distribution, radius, and endpoint grid alignment. The next section shows why changing only the codebook is insufficient; Section~\ref{sec:caq-zo} converts that failure mode into the aligned query rule.

\section{Weight-Space Queries Are Misaligned with Nonuniform Quantization Cells}
\label{sec:quantized-zo}

The companding view exposes a ZO-specific failure mode: the weight-space finite-difference radius \(\mu\) is chosen in \(x\), but rounding happens after the endpoints are mapped to the compander coordinate \(z=\phi(x)\). A weight-space stencil uses endpoints \(\vx\pm\mu\vu\), while the oracle evaluates the rounded endpoints \(Q(\vx\pm\mu\vu)\). Thus the first object to measure is not static reconstruction error, but how much of the \(z\)-grid the chosen stencil spans before rounding. A scalar coordinate slice \((x,u)\) gives this quantity. For \(Q=\phi^{-1}\circ U\circ\phi\), with \(U\) a uniform \(z\)-grid of spacing \(\Delta\), define
\begin{equation}
\label{eq:centered-grid-span}
\rho\defeq\frac{|\phi(x+\mu u)-\phi(x-\mu u)|}{2\Delta}.
\end{equation}
Here \(\rho\) is the normalized \(z\)-grid half-span of the scalar stencil \(x\pm\mu u\). It is defined by the two query endpoints and the grid spacing \(\Delta\). The following result shows that the same weight-space radius is converted into grid units by the local compander geometry.

\begin{theorem}[Fixed Weight-Space Radii Cannot Be Globally Grid-Calibrated]
\label{thm:no-global-radius}
Assume \(u\ne0\) and \(\phi\) is differentiable and strictly increasing on the interval between \(x-\mu u\) and \(x+\mu u\). Then for some \(\tilde x\) in this interval,
\begin{equation}
\rho
=
\frac{\mu |u|\nabla\phi(\tilde x)}{\Delta}.
\end{equation}
Thus the same weight-space finite-difference radius \(\mu\) is converted into grid units by the local compander slope \(\nabla\phi(\tilde x)\). In the local \(\mu\downarrow0\) sense, if \(\nabla\phi\) varies with location, no location-independent radius rule can keep \(\rho/(\mu|u|)\) calibrated across the domain; a location-independent infinitesimal match occurs only on regions where the compander is affine.
\end{theorem}

\begin{figure}[t]
\centering
\makebox[\textwidth][c]{\includegraphics[width=1.08\textwidth]{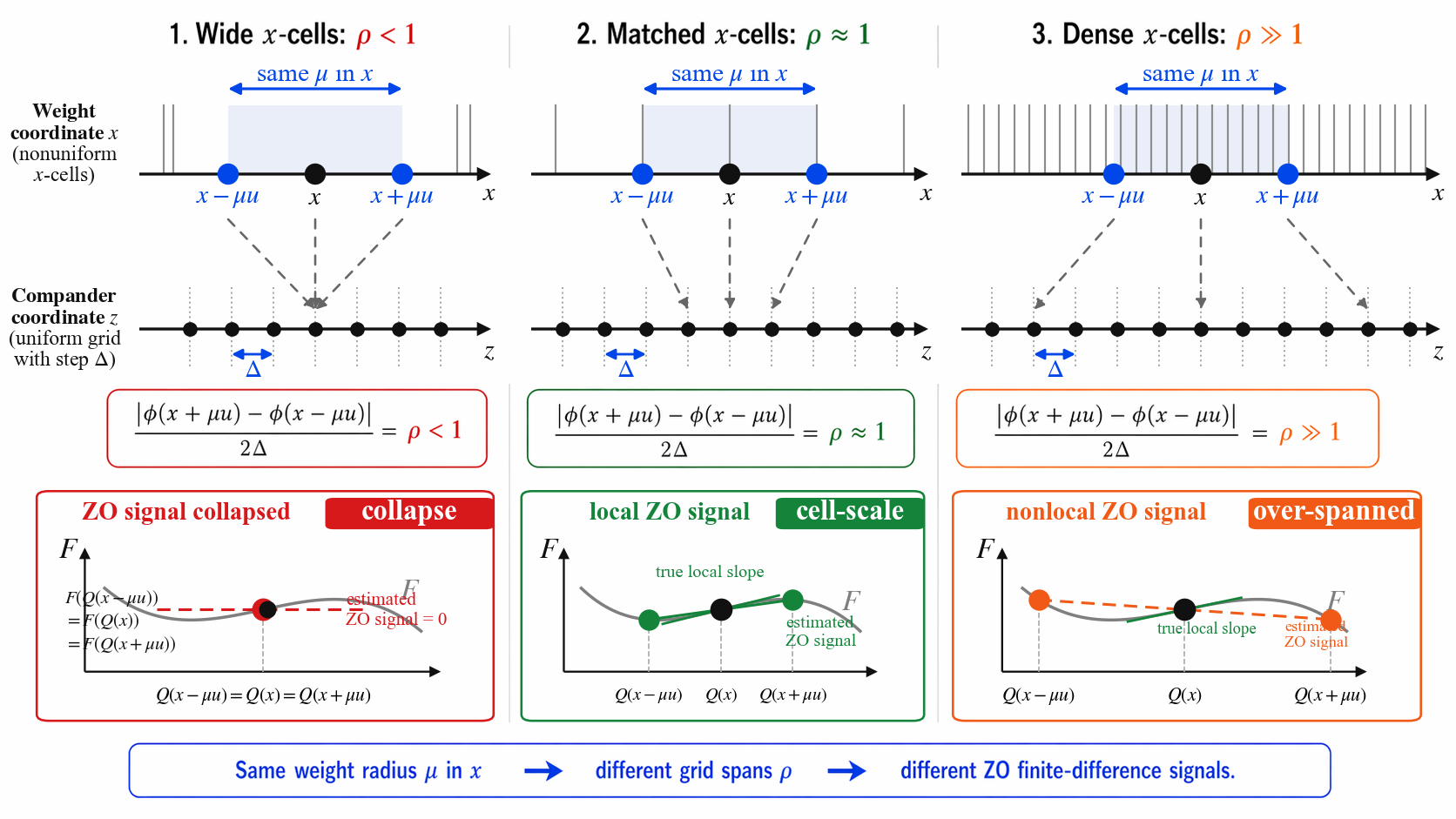}}
\caption{How grid-span mismatch becomes a ZO measurement distortion. The top and middle rows keep the same weight-space stencil \(x\pm\mu u\) and track its normalized span \(\rho\) on the uniform \(z\)-grid. The bottom row shows the corresponding finite-difference signal: same-cell collapse for \(\rho<1\), a local cell-scale response for \(\rho\simeq1\), and a nonlocal secant for \(\rho>1\).}
\label{fig:weight-space-zo-impact}
\end{figure}

This identity makes \(\nabla\phi(\tilde x)\) the local conversion factor from weight-space radius to grid units (proof in Appendix~\ref{app:no-global-radius-proof}). The resulting calibration mismatch produces three different ZO measurements. If \(\rho<1\), and especially if \(\rho\ll1\), the two rounded endpoints can fall in the same cell, so the measured estimate \(\hgrad(F\circ Q)(\vx;\vu)\) can collapse to zero. If \(\rho\simeq1\), the stencil is locally matched to the grid and the measured estimate is a local cell-scale secant. If \(\rho>1\), and especially if \(\rho\gg1\), the endpoints are resolved but the stencil over-spans the local grid scale, so the estimator measures a nonlocal secant and pays avoidable smoothing or curvature bias. Thus the grid spacing \(\Delta\) and the codebook induced by \(Q\) do not determine the ZO measurement by themselves; the stencil \(\{x-\mu u,x+\mu u\}\) relative to the \(U\)-cells in the \(z\)-coordinate also matters. Figure~\ref{fig:weight-space-zo-impact} visualizes this chain: a fixed weight-space radius becomes a location-dependent normalized \(z\)-grid span, and that span determines the finite-difference signal received by the optimizer.

The scalar signal distortion shown in Figure~\ref{fig:weight-space-zo-impact} becomes a vector-field issue for vector queries \((\vx,\vu)\). At the population level, the corresponding quantized two-point response is
\begin{equation}
\hgrad(F\circ Q)(\vx;\vu)
=
\frac{F(Q(\vx+\mu\vu))-F(Q(\vx-\mu\vu))}{2\mu}\,\vu .
\end{equation}
Its numerator is generated by the rounded chord \(Q(\vx+\mu\vu)-Q(\vx-\mu\vu)\), but the update is still placed in direction \(\vu\). In continuous ZO this chord is exactly \(2\mu\vu\); after weight-space quantized querying, the scalar distortions in Figure~\ref{fig:weight-space-zo-impact} can occur in different coordinates of the same vector query. The rounded chord is therefore a location-dependent vector and is generally not parallel to \(\vu\). Since descent depends on the alignment between \(\nabla F(\vx)\) and the expected estimator \(\E_{\vu}\!\left[\hgrad(F\circ Q)(\vx;\vu)\right]\), weight-space quantized ZO follows a location-dependent anisotropic field rather than the usual isotropic smoothing field. The endpoint-residual identity, rounded-chord field view, and standalone coupling bound are given in Appendices~\ref{app:proof-decomposition}, \ref{app:rounded-chord-field}, and~\ref{app:weight-space-coupling}. The design target is therefore to keep the measured chord aligned with the direction that receives the finite-difference response.

\section{CAQ-ZO: Compander-Aligned Queries}
\label{sec:caq-zo}

The preceding diagnosis isolates the mismatch that CAQ-ZO removes: in weight space, rounding changes the measured chord while the estimator still assigns the response to \(\vu\). CAQ-ZO instantiates the compander-aligned query geometry as an on-grid finite-difference rule for low-precision-forward-pass ZO, shown in Algorithm~\ref{alg:caq-zo}. It keeps the same quantizer \(Q=\phi^{-1}\circ U\circ\phi\) but forms the two query endpoints in the compander coordinate \(z=\phi(x)\), where \(U\) has uniform spacing \(\Delta\). By choosing endpoints that already lie on this grid, the measured chord and the update direction coincide in the coordinate where the finite difference is taken, rather than being separated by weight-space rounding.

CAQ-ZO combines two concrete mechanisms. First, it moves the stencil from weight space to the compander coordinate \(\vz=\phi(\vx)\), where \(U\) has a uniform grid with spacing \(\Delta\). Second, it uses a one-grid-step Rademacher stencil so that each queried coordinate remains exactly on that grid. Let \(\mathcal G\) be the \(z\)-grid of \(U\), assume the current iterate \(\vz_t\in\mathcal G\), and sample \(\vr\in\{-1,+1\}^d\). The query pair is \(\vz_t\pm\Delta\vr\). Within full-coordinate Rademacher stencils, one grid step is the smallest symmetric move that keeps every coordinate of both endpoints on the uniform \(z\)-grid. Consequently, away from clipping,
\begin{equation}
U(\vz_t\pm\Delta\vr)=\vz_t\pm\Delta\vr,\qquad
Q(\phi^{-1}(\vz_t\pm\Delta\vr))=\phi^{-1}(\vz_t\pm\Delta\vr).
\end{equation}
Thus CAQ-ZO keeps the same low-precision forward oracle and the same codebook, but changes the finite-difference stencil so that query-time quantization is the identity at the two endpoints.

With the on-grid stencil fixed, the update estimates the transformed-coordinate gradient directly. At iteration \(t\), CAQ-ZO samples \(\vr_1,\ldots,\vr_K\in\{-1,+1\}^d\) and queries the low-precision losses at \(\phi^{-1}(\vz_t+\Delta\vr_k)\) and \(\phi^{-1}(\vz_t-\Delta\vr_k)\). Because the endpoints are on-grid, the query-time quantizer contributes no additional residual; at the population level, the measured finite difference is a finite difference of \(F\circ\phi^{-1}\) at the queried points. With one stochastic sample \(\xi\) shared across the \(2K\) losses, the estimator is
\begin{equation}
\label{eq:caq-zo-estimator}
\hgrad_t(F\circ\phi^{-1})\defeq
\frac1K\sum_{k=1}^K
\frac{
f(\phi^{-1}(\vz_t+\Delta\vr_k);\xi)
-f(\phi^{-1}(\vz_t-\Delta\vr_k);\xi)
}{2\Delta}\,\vr_k
\end{equation}
where the equality uses the on-grid identity above. The algorithm then updates \(\vz\) directly using this sampled transformed-coordinate estimate, projects the stored iterate back to the \(z\)-grid, and maps back to \(\vx=\phi^{-1}(\vz)\) only to realize the quantized model state. This projection is an update-time constraint that prepares the next on-grid query pair; it is distinct from endpoint rounding during measurement. Weight-space queries carry that measurement channel, whereas CAQ-ZO removes it and pays only the optimization price of keeping the stored \(z\)-iterate on grid. Appendix~\ref{app:compressor-reduction} contrasts aligned and off-grid compander-coordinate querying; Appendices~\ref{app:query-time-residual-dichotomy}, \ref{app:estimator-error-proof}, \ref{app:on-grid-details}, \ref{app:uniform-grid-special-case}, and~\ref{app:algorithm} give the on-grid identity, estimator details, uniform-grid specialization, and implementation invariants.

\begin{algorithm}[t]
\caption{CAQ-ZO}
\label{alg:caq-zo}
\small
\begin{algorithmic}[1]
\STATE Initialize \(\vz_0=U(\phi(\vx_{\text{\normalfont init}}))\), \(\vx_0=\phi^{-1}(\vz_0)\), grid spacing \(\Delta\).
\FOR{\(t=0,\ldots,T-1\)}
  \STATE Sample \(\vr_1,\ldots,\vr_K\in\{-1,+1\}^d\) and one shared \(\xi\).
  \STATE Query losses at \(\phi^{-1}(\vz_t\pm\Delta\vr_k)\) for \(k=1,\ldots,K\).
  \STATE Form \(\hgrad_t(F\circ\phi^{-1})\) by Eq.~\eqref{eq:caq-zo-estimator}.
  \STATE Step and project: \(\vz_{t+1}=U(\vz_t-\eta\hgrad_t(F\circ\phi^{-1}))\), \(\vx_{t+1}=\phi^{-1}(\vz_{t+1})\).
\ENDFOR
\end{algorithmic}
\end{algorithm}

\section{Theoretical Guarantees and Comparisons}
\label{sec:theory}

Estimator error and descent expose the difference between compander alignment and a mere coordinate rewrite. The two estimators differ before any descent lemma is applied. A weight-space query forms \(\vx\pm\mu\vu\) and then asks the low-precision oracle to evaluate the rounded endpoints \(Q(\vx\pm\mu\vu)\). CAQ-ZO forms \(\vz\pm\Delta\vr\) directly on the uniform \(z\)-grid, where the endpoints are already fixed points of \(U\). This measurement order fixes the causal chain: endpoint rounding produces a residual, that residual propagates into finite-direction estimator error, and the estimator error determines the stationarity floor. All statements are local to the region visited by the algorithm and its probes, so \(\mathcal X\) and \(\mathcal Z\) denote compact neighborhoods containing the iterate path, query endpoints, rounded endpoints, and Taylor line segments. Full proofs appear in Appendix~\ref{app:sec5-full-proofs}, with auxiliary comparisons in Appendix~\ref{app:auxiliary-bounds}.

\begin{assumption}[Local Query Regularity]
\label{assump:local-regularity}
On \(\mathcal X\), \(F\) is \(L_x\)-smooth and lower bounded by \(F_\star\). On \(\mathcal Z\), \(F\circ\phi^{-1}\) is \(L_z\)-smooth and lower bounded by \((F\circ\phi^{-1})_\star\). Almost surely, \(f(\cdot;\xi)\) is \(L_x\)-smooth on \(\mathcal X\) and \(f(\phi^{-1}(\cdot);\xi)\) is \(L_z\)-smooth on \(\mathcal Z\). Differentiation and expectation commute on these local regions:
\[
\nabla F(\vx)=\E_{\xi}\!\left[\nabla f(\vx;\xi)\right],
\qquad
\nabla(F\circ\phi^{-1})(\vz)=\E_{\xi}\!\left[\nabla(f\circ\phi^{-1})(\vz;\xi)\right].
\]
\end{assumption}

\begin{assumption}[Interior Compander Grid and Expander Control]
\label{assump:interior-grid}
The uniform quantizer \(U\) acts on a \(z\)-grid \(\mathcal G\) with spacing \(\Delta\). Every \(z\)-point submitted to \(U\) remains in the unclipped round-to-nearest region, so \(U\) fixes grid points and moves any non-grid endpoint by at most \(\Delta/2\) per coordinate. The \(z\)-segment between each such point and its rounded value lies in \(\mathcal Z\). On \(\mathcal Z\), the expander is differentiable and satisfies \(\|\nabla\phi^{-1}(\vz)\|_{\mathrm{op}}\le B_\phi\).
\end{assumption}

These local assumptions separate the constants that enter the comparison. \(L_x\) and \(L_z\) control ordinary finite-difference smoothing in the two coordinates; \(B_\phi\) converts \(z\)-grid rounding into weight-space displacement; and the interior-grid condition makes CAQ-ZO endpoints exactly on-grid. For the stochastic objective from Section~\ref{sec:preliminaries}, \(F(\vx)=\E_{\xi}\!\left[f(\vx;\xi)\right]\), sampled-oracle fluctuations are tracked by local gradient-variance floors, as in stochastic ZO, rather than by a uniform sample-gradient envelope \citep{ghadimi2013zeroth,duchi2015zero}:
\begin{equation}
\begin{aligned}
\sigma_x^2
&\defeq \sup_{\vx\in\mathcal X}\E_{\xi}\!\left[\|\nabla f(\vx;\xi)-\nabla F(\vx)\|^2\right],\\
\sigma_z^2
&\defeq \sup_{\vz\in\mathcal Z}\E_{\xi}\!\left[\|\nabla(f\circ\phi^{-1})(\vz;\xi)-\nabla(F\circ\phi^{-1})(\vz)\|^2\right].
\end{aligned}
\label{eq:variance-profiles}
\end{equation}
These floors are coordinate-dependent: by the chain rule, the \(z\)-coordinate variance is shaped by the expander Jacobian and the anisotropy of the sampled-gradient covariance. CAQ-ZO removes the endpoint-rounding residual, not this ordinary coordinate-noise effect; the covariance accounting is deferred to Appendix~\ref{app:sampled-loss-bounds}. All bounds use common random numbers: at each iteration, one fresh \(\xi\) is shared across the \(2K\) endpoint evaluations. The notation keeps \(\xi\) only as the argument of sampled losses and lets expectations carry the oracle randomness. Throughout, \(A\lesssim B\) means \(A\le C_{\text{\normalfont num}}B\) for a universal numerical constant independent of \(d,K,T,\mu,\Delta,\eta\), the trajectory, and the quantizer except through displayed quantities, and \(c_\eta>0\) denotes a sufficiently small universal constant in step-size restrictions. The first result states the measurement residual that query alignment removes.

\begin{theorem}[Endpoint-Rounding Estimator Residual]
\label{thm:endpoint-rounding-response}
Under Assumptions~\ref{assump:local-regularity} and~\ref{assump:interior-grid}, let \(\vu_1,\ldots,\vu_K\) and \(\vr_1,\ldots,\vr_K\) be independent Rademacher directions, and let \(\hgrad(\cdot)\) without a direction argument denote the \(K\)-direction population response. For weight-space endpoints \(\vx\pm\mu\vu_k\),
\begin{equation}
\E\!\bigl[\|\hgrad(F\circ Q)(\vx)-\hgrad(F)(\vx)\|^2\bigr]\lesssim \frac{B_\phi^2d^2\Delta^2}{\mu^2}\left(\|\nabla F(\vx)\|^2+L_x^2d(\mu^2+B_\phi^2\Delta^2)\right),
\label{eq:weight-response-budget}
\end{equation}
whereas for CAQ-ZO endpoints \(\vz\pm\Delta\vr_k\), if \(\vz\in\mathcal G\), then
\begin{equation}
\label{eq:caq-response-zero}
\hgrad(F\circ Q\circ\phi^{-1})(\vz)-\hgrad(F\circ\phi^{-1})(\vz)=0 .
\end{equation}
\end{theorem}

Eq.~\eqref{eq:weight-response-budget} is the price of an off-grid stencil. A grid-scale displacement in \(z\) is expanded by \(\phi^{-1}\), producing the \(B_\phi\) factor, and then divided by the finite-difference radius \(\mu\). The bound keeps the current \(\|\nabla F(\vx)\|\) term and uses local \(L_x\)-smoothness only for nearby probed endpoints. CAQ-ZO instead uses \(\vz\pm\Delta\vr\), so \(U(\vz\pm\Delta\vr)=\vz\pm\Delta\vr\) in the unclipped interior and the measured quotient is already the continuous quotient. The estimator error is therefore standard ZO error plus the residual introduced by the query geometry.

\begin{theorem}[Estimator Error Under Query Geometry]
\label{thm:estimator-error-query-geometry}
Under Assumptions~\ref{assump:local-regularity} and~\ref{assump:interior-grid}, for independent Rademacher directions and sampled forward losses evaluated with one shared stochastic sample across the \(2K\) endpoint evaluations, the expectations include both the directions and the shared sample. The weight-space estimator satisfies
\begin{equation}
\begin{aligned}
\E\!\bigl[\|\hgrad(F\circ Q)(\vx)-\nabla F(\vx)\|^2\bigr]
&\lesssim \frac{d}{K}\|\nabla F(\vx)\|^2+L_x^2\mu^2d^3+\left(1+\frac{d}{K}\right)\sigma_x^2\\
&\quad+\frac{B_\phi^2d^2\Delta^2}{\mu^2}\left(\|\nabla F(\vx)\|^2+L_x^2d(\mu^2+B_\phi^2\Delta^2)+\sigma_x^2\right).
\end{aligned}
\label{eq:weight-estimator-budget}
\end{equation}
The CAQ-ZO estimator satisfies
\begin{equation}
\E\!\bigl[\|\hgrad(F\circ\phi^{-1})(\vz)-\nabla(F\circ\phi^{-1})(\vz)\|^2\bigr]\lesssim \frac{d}{K}\|\nabla(F\circ\phi^{-1})(\vz)\|^2+L_z^2\Delta^2d^3+\left(1+\frac{d}{K}\right)\sigma_z^2.
\label{eq:caq-estimator-budget}
\end{equation}
\end{theorem}

The first line of each estimator bound is standard stochastic Rademacher ZO. Increasing \(K\) reduces direction sampling, including the \(d\|\nabla F\|^2/K\) term and the direction-induced part of the sampled-loss contribution \citep{duchi2015zero,shamir2017optimal}. With one shared sample across the \(2K\) endpoints, the leading floor \(\sigma_x^2\) or \(\sigma_z^2\) remains as \(K\to\infty\); independent per-direction oracle samples would give a different variance term. Appendix~\ref{app:sampled-loss-bounds} derives the shared-sample bound and the coordinate-dependent variance floors. The bias terms \(L_x^2\mu^2d^3\) and \(L_z^2\Delta^2d^3\) use coordinatewise Rademacher steps, so the Euclidean radii are \(\mu\sqrt d\) and \(\Delta\sqrt d\). Other stochastic ZO regimes, including nonconvex stochastic estimators \citep{ghadimi2013zeroth}, two-point bandit estimators \citep{shamir2017optimal}, and higher-order estimators \citep{feng2023stochastic}, can change these factors.

The second line of Eq.~\eqref{eq:weight-estimator-budget} is the endpoint-rounding residual propagated to estimator error. Assumption~\ref{assump:interior-grid} gives \(\|Q(\vy)-\vy\|\le B_\phi\sqrt d\,\Delta/2\) at every probed weight point, and the quotient divides this displacement by \(\mu\). Local \(L_x\)-smoothness controls the sampled-gradient size along that segment by the current population gradient, the stochastic floor, and \(L_x^2d(\mu^2+B_\phi^2\Delta^2)\). Since \(\rho\approx \mu|\nabla\phi(y)u|/\Delta\) in one dimension, \(B_\phi^2d^2\Delta^2/\mu^2\) is the same under-resolution penalty as \(1/\rho^2\), up to local compander-slope variation. CAQ-ZO removes this radius trade-off by making grid alignment part of the stencil; with a deterministic oracle, the only remaining distinction is endpoint rounding versus zero.

\begin{theorem}[Stationarity Under Query Geometry]
\label{thm:caq-stationarity}
Under Assumptions~\ref{assump:local-regularity} and~\ref{assump:interior-grid}, consider the direct weight-space baseline
\(\vx_{t+1}=\vx_t-\eta\hgrad(F\circ Q)(\vx_t)\), with no storage projection, and suppose all iterates, probes, and rounded endpoints remain in \(\mathcal X\). There exist universal numerical constants \(c_{\text{\normalfont res}},c_\eta>0\) such that, if
\(B_\phi^2d^2\Delta^2/\mu^2\le c_{\text{\normalfont res}}\) and \(\eta\le c_\eta/[L_x(1+d/K+B_\phi^2d^2\Delta^2/\mu^2)]\), then
\begin{equation}
\begin{aligned}
\frac1T\sum_{t=0}^{T-1}\E\!\bigl[\|\nabla F(\vx_t)\|^2\bigr]
&\lesssim \frac{F(\vx_0)-F_\star}{\eta T}
+L_x^2\mu^2d^3
+L_x\eta\left(1+\frac{d}{K}+\frac{B_\phi^2d^2\Delta^2}{\mu^2}\right)\sigma_x^2\\
&\quad+\frac{B_\phi^2d^2\Delta^2}{\mu^2}L_x^2d(\mu^2+B_\phi^2\Delta^2).
\end{aligned}
\label{eq:weight-space-stationarity-budget}
\end{equation}
For CAQ-ZO, suppose the iterates follow Algorithm~\ref{alg:caq-zo}, with \(\vx_t=\phi^{-1}(\vz_t)\), sampled responses \(\vz_{t+1}=U(\vz_t-\eta\hgrad_t(F\circ\phi^{-1}))\), and pre-projection points \(\bar{\vz}_{t+1}\defeq \vz_t-\eta\hgrad_t(F\circ\phi^{-1})\). Assume the query endpoints, pre-projection points, projected iterates, and connecting \(z\)-segments remain in the unclipped round-to-nearest region and in \(\mathcal Z\), so \(\|U(\bar{\vz}_{t+1})-\bar{\vz}_{t+1}\|_\infty\le\Delta/2\). If
\(\eta\le c_\eta/[L_z(1+d/K)]\), then
\begin{equation}
\frac1T\sum_{t=0}^{T-1}\E\!\bigl[\|\nabla(F\circ\phi^{-1})(\vz_t)\|^2\bigr]
\lesssim \frac{F(\vx_0)-(F\circ\phi^{-1})_\star}{\eta T}+L_z^2\Delta^2d^3+L_z\eta\left(1+\frac{d}{K}\right)\sigma_z^2+\frac{d\Delta^2}{\eta^2},
\label{eq:caq-stationarity-budget}
\end{equation}
\end{theorem}

The stationarity bound turns the estimator residual into a descent condition. In the weight-space baseline, \(B_\phi^2d^2\Delta^2/\mu^2\) is both a second-moment inflation, handled by the step-size restriction, and a conditional-bias source, which requires the small-alignment condition and produces the residual floor. CAQ-ZO has no query-time alignment condition because Theorem~\ref{thm:endpoint-rounding-response} makes the measurement residual zero; it keeps only finite-difference bias, \(L\eta\)-scaled sampled-loss variance, and the storage-projection term \(d\Delta^2/\eta^2\). This last term is absent from the direct baseline because \(\vx_t\) is stored continuously; a grid-projected weight-space variant would add its own storage-grid term. Appendix~\ref{app:transformed-convergence} gives the descent calculation behind this comparison.

\begin{theorem}[Weight-Space Stationarity Transfer]
\label{thm:caq-weight-space-stationarity}
Under the conditions of Theorem~\ref{thm:caq-stationarity}, suppose additionally that
\(\nabla\phi^{-1}(\vz_t)\nabla\phi^{-1}(\vz_t)^\top\succeq m_\phi\mI\) for all iterates \(t\), with \(m_\phi>0\), where \(\nabla\phi^{-1}\) denotes the expander Jacobian. If \(\vx_t=\phi^{-1}(\vz_t)\), then
\begin{equation}
\frac1T\sum_{t=0}^{T-1}\E\!\bigl[\|\nabla F(\vx_t)\|^2\bigr]
\lesssim \frac{1}{m_\phi}\!\left[\frac{F(\vx_0)-(F\circ\phi^{-1})_\star}{\eta T}+L_z^2\Delta^2d^3+L_z\eta\left(1+\frac{d}{K}\right)\sigma_z^2+\frac{d\Delta^2}{\eta^2}\right].
\label{eq:caq-weight-space-transfer}
\end{equation}
\end{theorem}

The transfer theorem is purely a conditioning statement. The lower singular-value condition is what lets a \(z\)-stationarity certificate imply a weight-space one; without it, \(\|\nabla(F\circ\phi^{-1})(\vz)\|\) may be small while \(\|\nabla F(\vx)\|\) is not, independently of quantization. Appendix~\ref{app:compressor-smoothness-cost} expands the coordinate-conditioning calculation. The comparison is therefore governed by two axes: query-time coupling, which creates the endpoint-rounding floor and the resolution condition for weight-space descent, and coordinate noise, which enters through the separate floors \(\sigma_x^2\) and \(\sigma_z^2\). Section~\ref{sec:experiments} targets the regime singled out by these bounds: compander alignment becomes visible when endpoint rounding dominates ordinary finite-difference and stochastic-oracle errors.

\section{Experiments}
\label{sec:experiments}

The experiments test the query-geometry claim at two levels. The synthetic comparison fixes companders, starts, direction count, optimizer state, scale calibration, and endpoint rounding, then changes only the finite-difference stencil. The LLM comparison fixes the NF4 codebook, bit-width, model, task budget, and validation split, then compares QuZO weight-space queries with CAQ-ZO compander-aligned queries. Accordingly, the main evidence consists of the synthetic convergence figure and the matched NF4 LLM table; Appendix~\ref{app:experiments} contains the estimator-residual measurement and the LLM search grid.

\subsection{Synthetic Mechanism Validation}
\label{sec:synthetic-validation}

The synthetic setting makes endpoint rounding visible by using few directions in high dimension. We compare \(\mu\)-law 2-bit and NF4-style 4-bit companders on Quadratic, Levy, Rosenbrock, and Ackley objectives at \(d=10000\), \(K=4\), \(T=10000\), and three start-matched initializations. All methods share per-block scale calibration, dynamic scale recalibration, an FP master state, Adam, clipping, deterministic endpoint rounding, and \(\eta=0.005\); no hyperparameter is tuned per method, objective, or quantizer. CAQ-ZO queries on-grid Rademacher endpoints in the compander coordinate, whereas Vanilla Gaussian ZO and QuZO query weight-space endpoints before the same quantized forward evaluation.

\begin{figure}[t]
\centering
\setlength{\floatsep}{4pt plus 1pt minus 1pt}
\setlength{\textfloatsep}{6pt plus 1pt minus 2pt}
\includegraphics[trim=3pt 3pt 3pt 5pt,clip,width=\textwidth]{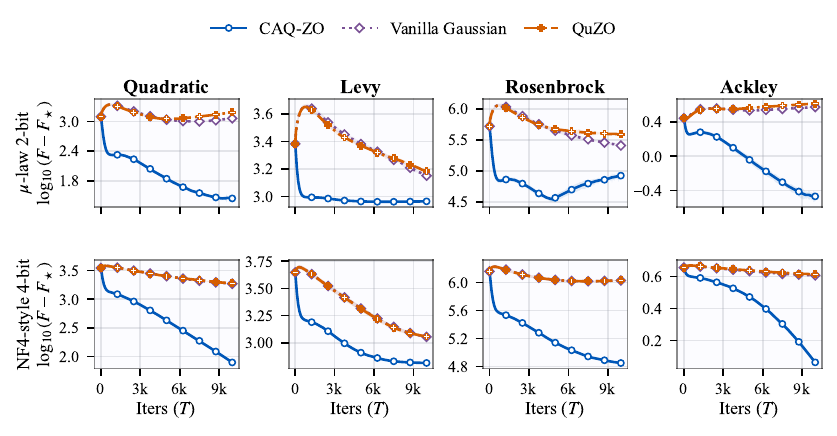}
\setlength{\abovecaptionskip}{2pt}
\setlength{\belowcaptionskip}{2pt}
\caption{Start-matched synthetic convergence under matched nonuniform low-bit forward evaluation. Rows are \(\mu\)-law 2-bit and NF4-style 4-bit; columns are four objectives at \(d=10000\). All curves use \(K=4\), \(T=10000\), three starts, the same FP-master/Adam update, deterministic endpoint rounding, and \(\eta=0.005\). CAQ-ZO uses on-grid compander stencils; baselines use weight-space stencils.}
\label{fig:main-synthetic-convergence}
\end{figure}

Appendix~\ref{app:experiments} measures the predicted error term by subtracting the unrounded estimator with identical directions: CAQ-ZO stays at the numerical floor, while weight-space baselines retain the endpoint-rounding residual from Theorems~\ref{thm:endpoint-rounding-response} and~\ref{thm:estimator-error-query-geometry}. Figure~\ref{fig:main-synthetic-convergence} shows the corresponding optimization effect. CAQ-ZO has the lowest final/start gap ratio in all eight panels, with the clearest separation on Quadratic and Rosenbrock; Levy and Ackley add landscape and finite-direction effects, but preserve the ordering. The pattern matches the stationarity comparison in Section~\ref{sec:theory}: once the quantized forward channel creates a persistent endpoint-rounding term for weight-space stencils, increasing iterations alone does not remove the induced vector-field distortion. CAQ-ZO changes the measurement rather than the optimizer wrapper, so its advantage appears under the same Adam state, scale recalibration, and query budget.

\subsection{LLM Empirical Evidence}
\label{sec:llm-empirical}

The LLM study fine-tunes Qwen-2.5-1.5B \citep{qwen2025qwen25} and Llama-2-7B \citep{touvron2023llama2} through NF4 forward evaluation \citep{dettmers2023qlora}. The trained NF4 comparison is QuZO weight-space queries \citep{zhou2025quzo} versus CAQ-ZO compander-aligned queries under the same model, codebook, bit-width, task splits, search grid, and budget. Zero-shot NF4 and BF16 MeZO \citep{malladi2023mezo} give quantized-start/high-precision references; six tasks use 1{,}000/500/1{,}000 train/validation/test splits, with metrics and grids in Appendix~\ref{app:experiments}.

\begin{table}[t]
\centering
\setlength{\abovecaptionskip}{2pt}
\setlength{\belowcaptionskip}{2pt}
\caption{LLM results under NF4 forward evaluation. MeZO (BF16) is separated as a high-precision reference; all NF4 rows share the compander. \textbf{Bold} marks the better trained NF4 method.}
\label{tab:llm-finetune}
\small
\setlength{\tabcolsep}{3.3pt}
\renewcommand{\arraystretch}{0.90}
\begin{tabular}{@{}llcccccc@{}}
\toprule
 & & \multicolumn{4}{c}{Classification} & \multicolumn{1}{c}{Mult.-Choice} & \multicolumn{1}{c}{Generation} \\
\cmidrule(lr){3-6}\cmidrule(lr){7-7}\cmidrule(lr){8-8}
Model & Method & SST-2 & RTE & BoolQ & CB & COPA & SQuAD \\
\midrule
Qwen-2.5-1.5B & MeZO (BF16)       & 80.2 & 72.1 & 65.4 & 69.5 & 75.1 & 38.8 \\
\cmidrule(l){2-8}
 & Zero-Shot-Q (NF4) & 60.1 & 53.8 & 65.8 & 17.8 & 73.0 & 19.1 \\
 & QuZO (NF4)        & 50.9 & 52.6 & 61.3 & 41.7 & 63.9 &  9.3 \\
 & \textbf{CAQ-ZO (NF4)} & \textbf{71.3} & \textbf{61.5} & \textbf{62.3} & \textbf{60.5} & \textbf{76.2} & \textbf{33.3} \\
\midrule
Llama-2-7B & MeZO (BF16)       & 83.5 & 58.1 & 69.6 & 67.9 & 78.0 & 80.7 \\
\cmidrule(l){2-8}
 & Zero-Shot-Q (NF4) & 57.2 & 55.6 & 57.2 & 36.4 & 78.0 & 53.6 \\
 & QuZO (NF4)        & 72.1 & 56.5 & 64.9 & 37.3 & 77.9 & 30.5 \\
 & \textbf{CAQ-ZO (NF4)} & \textbf{76.5} & \textbf{57.4} & \textbf{67.2} & \textbf{58.7} & \textbf{78.4} & \textbf{58.6} \\
\bottomrule
\end{tabular}
\end{table}

Table~\ref{tab:llm-finetune} gives the trained NF4 comparison under matched splits, search grid, quantizer, and budget. CAQ-ZO improves over QuZO on every metric for both models; the largest gains are CB/SQuAD (\(18.8/21.4\) accuracy points and \(24.0/28.1\) F1 for Qwen/Llama), and the rest are smaller but positive. BF16 MeZO is a high-precision reference rather than a matched quantized baseline. The matched NF4 result therefore supports the query-geometry claim in the intended low-precision-forward-pass setting: changing the finite-difference endpoints, while holding the codebook fixed, improves the trained quantized model.

\section{Conclusion}

Quantized ZO is not determined solely by the codebook: low-precision forward passes also measure finite-difference stencils. Nonuniform cells can turn weight-space stencils into rounded chords; CAQ-ZO removes this channel by querying one-grid-step Rademacher endpoints in the compander coordinate. The theory gives estimator and stationarity comparisons, and the experiments verify synthetic residuals and matched NF4 LLM gains. The scope is scalar companding quantizers with scoped NF4 evidence; non-scalar codebooks and broader large-model claims need separate validation.

\bibliographystyle{plainnat}
\bibliography{workspace/reference}

\appendix
\section{Related Work}
\label{app:related-work}

Classical derivative-free and simultaneous-perturbation methods estimate gradients from function values rather than backpropagated derivatives \citep{spall1992spsa,nesterov2017random,duchi2015zero,ghadimi2013zeroth}. Recent large-model work revives this line because forward-only optimization can substantially reduce memory in language-model fine-tuning; MeZO adapts ZO-SGD to operate in place and shows that prompt-conditioned pretrained models can be fine-tuned with inference-level memory \citep{malladi2023mezo}. Subsequent work improves variance, masking, and temporal structure, but the usual estimator analysis treats the queried function as continuous or only externally noisy. CAQ-ZO studies the missing case where the function evaluation is passed through a low-bit quantizer and the estimator error depends on how ZO perturbations interact with the quantizer's cells.

Recent work combines quantization and ZO under several distinct targets. Fixed-point forward-gradient and MCU training work asks whether inference-style low-precision hardware can support on-device adaptation without backpropagation \citep{feng2024stepping,zhao2024poor}. QuZO fine-tunes LLMs through low-bit forward passes using quantized perturbations and stochastic rounding \citep{zhou2025quzo}. QZO avoids perturbing discrete weights by perturbing continuous quantization scales, making it compatible with scalar and codebook PTQ but changing the optimized variables from weights to scales \citep{shang2026qzo}. ZOQO and QES move closer to discrete-space optimization by updating quantized parameters directly, using sign-based quantized noise or accumulated error feedback to keep small update signals from vanishing on the integer lattice \citep{bar2025zoqo,xu2026qes}. Quantized-model test-time adaptation uses ZO to adapt deployed quantized models under domain shift, but its focus is streaming robustness and selected adaptation variables rather than finite-difference query geometry \citep{deng2025testtime}. CAQ-ZO is closest in setting to the low-precision-forward-pass branch, especially QuZO and QZO-FF, and closest in discrete-query motivation to ZOQO and QES. The distinct object here is query geometry: for a fixed companding-representable codebook, which finite-difference stencil makes measurement matched to the quantizer?

Scalar quantization theory has long distinguished uniform quantization, Lloyd--Max codebooks, and companding transformations that map a nonuniform source through a compressor before uniform quantization \citep{max1960quantizing,lloyd1982least,gray1998quantization}. Modern neural quantization uses related ideas in several forms: integer-arithmetic QAT \citep{jacob2018quantization}, LLM post-training quantization such as GPTQ and AWQ \citep{frantar2022gptq,lin2023awq}, and low-bit adaptation methods such as QLoRA and LoftQ \citep{dettmers2023qlora,li2024loftq}. These methods primarily optimize storage, inference accuracy, or backpropagation-through-quantization workflows. The present focus is orthogonal: even when the codebook is fixed, ZO faces the additional choice of whether its perturbations are aligned with the compander. This explains why a codebook that is favorable for static reconstruction can be unfavorable for weight-space perturbations.

Quantization-aware training usually simulates low-bit arithmetic during differentiable training and uses surrogate gradients or low-rank adaptation to recover accuracy \citep{jacob2018quantization,liu2024lrqat,wang2024efficientqat}. Those methods can rely on backpropagation and therefore do not need to choose how random perturbations interact with quantization cells for each function-value evaluation. In quantization-aware ZO, the query itself is the measurement instrument. CAQ-ZO identifies compander-aligned queries as the query-geometry-matched design and uses the corresponding on-grid Rademacher stencil in the compander coordinate. Thus the novelty is not a new nonuniform codebook; it is a compander-aligned query geometry for ZO estimators under an existing companding-representable quantizer.

\section{Mechanism and Main Proof Details}
\label{app:proofs}

\subsection{Endpoint-Residual Identity}
\label{app:proof-decomposition}

The coupling gap between quantized and unquantized responses is defined for the same weight-space stencil. Define
\(\varepsilon_\pm\defeq F(Q(\vx\pm\mu\vu))-F(\vx\pm\mu\vu)\). Subtracting the unquantized two-point response from the quantized two-point response gives the endpoint-residual expansion
\begin{equation}
\label{eq:coupling-term}
\hgrad(F\circ Q)(\vx;\vu)-\hgrad(F)(\vx;\vu)
=
\frac{\varepsilon_+-\varepsilon_-}{2\mu}\vu .
\end{equation}
No stochastic rounding, unbiased residual model, or smoothness assumption is needed for this population-response identity. Smoothness and stochastic finite-difference noise enter later when bounding the size of the estimator error terms.

For a compander-coordinate finite-difference stencil, the quantized evaluation map is
\begin{equation}
Q(\phi^{-1}(\vz\pm\mu\vu))=\phi^{-1}(U(\vz\pm\mu\vu)).
\end{equation}
With \(H\defeq F\circ\phi^{-1}\), this gives the compander-coordinate coupling gap
\begin{equation}
\hgrad(H\circ U)(\vz;\vu)-\hgrad H(\vz;\vu).
\end{equation}
This form is the basis for both the off-grid compander-coordinate residual bound and the CAQ-ZO on-grid query identity.

\subsection{Rounded Chords and the Weight-Space ZO Field}
\label{app:rounded-chord-field}

The rounded-chord view in Section~\ref{sec:quantized-zo} can be written as an exact identity before any local approximation is made. Let
\begin{equation}
\tilde{\vx}_\pm\defeq Q(\vx\pm\mu\vu),
\qquad
\va_Q(\vx,\vu;\mu)\defeq
\frac{\tilde{\vx}_+-\tilde{\vx}_-}{2\mu}.
\end{equation}
If \(F\) is differentiable on the segment between the rounded endpoints, the fundamental theorem of calculus gives
\begin{equation}
F(\tilde{\vx}_+)-F(\tilde{\vx}_-)
=
(\tilde{\vx}_+-\tilde{\vx}_-)^\top
\bar{\vg}(\vx,\vu;\mu),
\qquad
\bar{\vg}(\vx,\vu;\mu)
\defeq
\int_0^1
\nabla F\!\left(\tilde{\vx}_-+s(\tilde{\vx}_+-\tilde{\vx}_-)\right)\,ds .
\end{equation}
Therefore the quantized two-point response satisfies
\begin{equation}
\hgrad(F\circ Q)(\vx;\vu)
=
\left(\va_Q(\vx,\vu;\mu)^\top\bar{\vg}(\vx,\vu;\mu)\right)\vu .
\end{equation}
Without quantization, \(\va_Q=\vu\), so the response is the usual projection onto the same direction used for the update. With weight-space quantized querying, the response is measured along \(\va_Q\) but assigned to \(\vu\).

For an \(L\)-smooth reference loss and an update \(\vx^+=\vx-\eta\hgrad(F\circ Q)(\vx;\vu)\), the standard descent lemma gives
\begin{equation}
\E\!\left[F(\vx^+)\mid\vx\right]
\le
F(\vx)
-\eta\left\langle\nabla F(\vx),\E\!\left[\hgrad(F\circ Q)(\vx;\vu)\mid\vx\right]\right\rangle
+\frac{L\eta^2}{2}\E\!\left[\|\hgrad(F\circ Q)(\vx;\vu)\|^2\right] .
\end{equation}
Thus the descent-relevant quantity is the alignment between \(\nabla F(\vx)\) and the expected quantized ZO field. A local linearization of the exact rounded-chord identity gives the response form
\begin{equation}
\E\!\left[\hgrad(F\circ Q)(\vx;\vu)\mid\vx\right]
\approx
\mM_Q(\vx,\mu)\nabla F(\vx),
\qquad
\mM_Q(\vx,\mu)\defeq
\E_{\vu}\!\left[\vu\,\va_Q(\vx,\vu;\mu)^\top\right].
\end{equation}
For continuous isotropic ZO, this map reduces to the identity up to ordinary smoothing error. Under weight-space quantized querying, it is a location-dependent, quantizer-induced map: under-resolved cells shrink descent-relevant directions, while over-spanned coordinates transmit nonlocal secants with larger smoothing bias. Heterogeneous coordinate regimes make the resulting field anisotropic rather than a scalar rescaling of the usual ZO field.

\subsection{Proof of the Grid-Span Theorem}
\label{app:no-global-radius-proof}

The theorem is a scalar coordinate slice of the vector query. The mean-value theorem gives
\begin{equation}
\phi(x+\mu u)-\phi(x-\mu u)
=
2\mu u\nabla\phi(\tilde x)
\end{equation}
for some \(\tilde x\) between \(x-\mu u\) and \(x+\mu u\). Since the theorem assumes \(\phi\) is strictly increasing on the interval, this gives
\begin{equation}
\rho
=
\frac{|\phi(x+\mu u)-\phi(x-\mu u)|}{2\Delta}
=
\frac{\mu |u|\nabla\phi(\tilde x)}{\Delta}.
\end{equation}
Dividing by \(\mu |u|\) and taking \(\mu\downarrow0\) yields the local compander-grid span limit
\begin{equation}
\lim_{\mu\downarrow0}\frac{\rho}{\mu |u|}
=
\frac{\nabla\phi(x)}{\Delta}.
\end{equation}
If \(\nabla\phi\) is nonconstant on the scalar coordinate domain, two locations have different infinitesimal grid-scale conversion factors. Matching a fixed local normalized conversion \(c\) at \(x\) would require \(\mu=c\Delta/(|u|\nabla\phi(x))\) to first order, hence a location-dependent radius. Conversely, a location-independent infinitesimal grid-span conversion over an interval requires \(\nabla\phi\) to be constant on that interval, equivalently \(\phi\) is affine there. The theorem uses this infinitesimal conclusion and does not assert a finite-window converse.

\subsection{Proofs for Section~\ref{sec:theory}}
\label{app:sec5-full-proofs}

\paragraph{Universal constants.}
The symbol \(C\) is not a new problem parameter. The theorem statements display every problem-dependent quantity explicitly and absorb only universal numerical multipliers through \(\lesssim\). These multipliers come from elementary inequalities used repeatedly below:
\begin{equation}
\|\va+\vb\|^2\le 2\|\va\|^2+2\|\vb\|^2,\qquad
\langle \va,\vb\rangle\le \frac{\gamma}{2}\|\va\|^2+\frac{1}{2\gamma}\|\vb\|^2,
\qquad
\left\|\frac1K\sum_{k=1}^K\va_k\right\|^2\le \frac1K\sum_{k=1}^K\|\va_k\|^2 .
\end{equation}
They also absorb fixed numerical factors from the smoothness-integral remainder and from the Rademacher identity \(\E\!\left[\vr\vr^\top\right]=\mI\). Tracking constants such as \(1/4,2,6,16\) would not change any displayed dependency. The proof keeps all dimension, radius, smoothness, variance, expander, sampled-oracle, and projection dependencies visible, and uses \(C\) only for universal numerical multipliers independent of \(d,K,T,\mu,\Delta,\eta\), the trajectory, and the quantizer beyond the constants named in Assumptions~\ref{assump:local-regularity} and~\ref{assump:interior-grid}.

\subsubsection{Proof of Theorem~\ref{thm:endpoint-rounding-response}}
\label{app:query-time-residual-dichotomy}

The proof covers the two cases in Theorem~\ref{thm:endpoint-rounding-response} and states the non-grid compander-coordinate residual used in the auxiliary comparisons below. Throughout, \(\|\vr\|=\|\vu\|=\sqrt d\) for Rademacher directions.

\paragraph{Non-grid compander-coordinate residual.}
Let \(H\defeq F\circ\phi^{-1}\) and define the endpoint rounding errors
\begin{equation}
\ve_\pm\defeq U(\vz\pm\mu\vr)-(\vz\pm\mu\vr).
\end{equation}
Assumption~\ref{assump:interior-grid} gives \(\|\ve_\pm\|_\infty\le \Delta/2\), hence
\begin{equation}
\|\ve_\pm\|\le \frac{\sqrt d\,\Delta}{2}.
\label{eq:rounding-residual-norm}
\end{equation}
For one direction,
\begin{equation}
\begin{aligned}
&\hgrad(F\circ Q\circ\phi^{-1})(\vz;\vr)-\hgrad(F\circ\phi^{-1})(\vz;\vr)\\
&\quad =
\frac{
H(\vz+\mu\vr+\ve_+)-H(\vz+\mu\vr)
-
H(\vz-\mu\vr+\ve_-)+H(\vz-\mu\vr)
}{2\mu}\,\vr .
\end{aligned}
\end{equation}
Any point on the segment between \(\vz\pm\mu\vr\) and \(\vz\pm\mu\vr+\ve_\pm\) is within distance \(\mu\sqrt d+\sqrt d\,\Delta/2\) of \(\vz\). By \(L_z\)-smoothness of \(H\), the integral form of the fundamental theorem of calculus gives
\begin{equation}
\begin{aligned}
\left\|\hgrad(F\circ Q\circ\phi^{-1})(\vz;\vr)-\hgrad(F\circ\phi^{-1})(\vz;\vr)\right\|^2
\lesssim \frac{d^2\Delta^2}{\mu^2}
\left(\|\nabla H(\vz)\|^2+L_z^2d(\mu^2+\Delta^2)\right).
\end{aligned}
\end{equation}
For the \(K\)-direction average, let \(\vb_k\) denote the left-hand residual above with direction \(\vr_k\). Jensen's inequality gives
\begin{equation}
\E\!\left[\left\|\frac1K\sum_{k=1}^K\vb_k\right\|^2\right]\lesssim \frac{d^2\Delta^2}{\mu^2}\left(\|\nabla H(\vz)\|^2+L_z^2d(\mu^2+\Delta^2)\right).
\label{eq:offgrid-response-budget}
\end{equation}
This gives the residual response budget under the same local smoothness convention as the theorem statements.

\paragraph{Weight-space queries.}
For one direction, Eq.~\eqref{eq:coupling-term} gives the exact endpoint-residual identity. Since \(U\) moves a non-grid \(z\)-endpoint by at most \(\sqrt d\,\Delta/2\), the corresponding \(z\)-segment remains in \(\mathcal Z\), and \(\|\nabla\phi^{-1}\|_{\mathrm{op}}\le B_\phi\) on \(\mathcal Z\),
\begin{equation}
\|Q(\vx\pm\mu\vu)-(\vx\pm\mu\vu)\|\le \frac{B_\phi\sqrt d\,\Delta}{2}.
\end{equation}
Any point on the segment between \(\vx\pm\mu\vu\) and \(Q(\vx\pm\mu\vu)\) is within distance \(\mu\sqrt d+B_\phi\sqrt d\,\Delta/2\) of \(\vx\). By \(L_x\)-smoothness,
\begin{equation}
\|\nabla F(\tilde\vx)\|^2\lesssim \|\nabla F(\vx)\|^2+L_x^2d(\mu^2+B_\phi^2\Delta^2)
\end{equation}
along this segment. The integral form of the fundamental theorem of calculus gives
\begin{equation}
\|\hgrad(F\circ Q)(\vx;\vu)-\hgrad(F)(\vx;\vu)\|^2\lesssim \frac{B_\phi^2d^2\Delta^2}{\mu^2}\left(\|\nabla F(\vx)\|^2+L_x^2d(\mu^2+B_\phi^2\Delta^2)\right).
\end{equation}
Jensen's inequality over the \(K\) directions proves Eq.~\eqref{eq:weight-response-budget}.

\paragraph{CAQ-ZO queries.}
If \(\vz\in\mathcal G\) and the interior condition holds, then the Rademacher one-grid-step endpoints remain grid points:
\begin{equation}
\vz\pm\Delta\vr\in\mathcal G,\qquad U(\vz\pm\Delta\vr)=\vz\pm\Delta\vr.
\end{equation}
Consequently,
\begin{equation}
Q(\phi^{-1}(\vz\pm\Delta\vr))
=\phi^{-1}(U(\vz\pm\Delta\vr))
=\phi^{-1}(\vz\pm\Delta\vr).
\end{equation}
The two measured endpoint losses are exactly the continuous losses of \(F\circ\phi^{-1}\), so
\begin{equation}
\hgrad(F\circ Q\circ\phi^{-1})(\vz;\vr)=\hgrad(F\circ\phi^{-1})(\vz;\vr).
\end{equation}
Averaging over \(K\) directions gives Eq.~\eqref{eq:caq-response-zero}.

\subsubsection{Proof of Theorem~\ref{thm:estimator-error-query-geometry}}
\label{app:estimator-error-proof}

The estimator bound combines the ordinary continuous Rademacher finite-difference error, the endpoint-rounding residual just proved, and a sampled-loss upper bound for the measured estimator.

\paragraph{Continuous Rademacher finite-difference error.}
Let \(G\) be an \(L_G\)-smooth scalar objective on a local region, and let \(\vp\defeq\nabla G(\vy)\). For radius \(\delta\) and \(\vr\in\{-1,+1\}^d\), define
\begin{equation}
\hgrad G(\vy;\vr)\defeq\frac{G(\vy+\delta\vr)-G(\vy-\delta\vr)}{2\delta}\,\vr .
\end{equation}
The integral form of the fundamental theorem of calculus gives
\begin{equation}
\frac{G(\vy+\delta\vr)-G(\vy-\delta\vr)}{2\delta}
=\frac{1}{2\delta}\int_{-\delta}^{\delta}\langle\nabla G(\vy+s\vr),\vr\rangle\,ds .
\end{equation}
Hence
\begin{equation}
\hgrad G(\vy;\vr)=\langle\vp,\vr\rangle\vr+\tau_{\vr},
\qquad
\|\tau_{\vr}\|\le \frac{L_G\delta}{2}\|\vr\|^3
=\frac{L_G\delta d^{3/2}}{2}.
\label{eq:smooth-rademacher-remainder}
\end{equation}
Rademacher orthogonality gives
\begin{equation}
\E\!\left[\langle \vp,\vr\rangle\vr\right]=\E\!\left[\vr\vr^\top\right]\vp=\vp,
\end{equation}
and
\begin{equation}
\begin{aligned}
\E\!\left[\|\langle \vp,\vr\rangle\vr-\vp\|^2\right]
&=\E\!\left[\|\langle \vp,\vr\rangle\vr\|^2\right]-\|\vp\|^2\\
&=d\,\E\!\left[\langle \vp,\vr\rangle^2\right]-\|\vp\|^2
=(d-1)\|\vp\|^2 .
\end{aligned}
\label{eq:rademacher-sampling-variance}
\end{equation}
For \(K\) independent directions, combine Eqs.~\eqref{eq:smooth-rademacher-remainder}--\eqref{eq:rademacher-sampling-variance} with \(\|\va+\vb\|^2\le 2\|\va\|^2+2\|\vb\|^2\) and Jensen's inequality:
\begin{equation}
\E\!\left[
\left\|
\frac1K\sum_{k=1}^K\hgrad G(\vy;\vr_k)-\nabla G(\vy)
\right\|^2
\right]
\le
C\frac{d}{K}\|\nabla G(\vy)\|^2
+CL_G^2\delta^2d^3 .
\label{eq:continuous-rademacher-budget}
\end{equation}
The constant \(C\) here only absorbs fixed numerical factors.
This is the coordinatewise Rademacher analogue of the random-direction moment calculations used in standard two-point ZO analyses \citep{duchi2015zero,ghadimi2013zeroth}; normalized sphere or orthogonal-direction variants follow different moment identities \citep{shamir2017optimal,feng2023stochastic}.

\paragraph{Sampled-loss upper bounds.}
\label{app:sampled-loss-bounds}
Population smoothness alone cannot control sampled losses: adding a zero-mean sample-dependent function to \(F\) can leave \(F\) unchanged while changing sampled endpoint values. The sampled-oracle contribution is therefore represented by the domain-level variance scalars \(\sigma_x^2\) and \(\sigma_z^2\), rather than by a uniform gradient-norm envelope.

The coordinate relation between these two scalars is as follows. Let
\begin{equation}
J(\vz)\defeq\nabla\phi^{-1}(\vz),\qquad
\delta_x(\vx;\xi)\defeq\nabla f(\vx;\xi)-\nabla F(\vx).
\end{equation}
For \(\vz=\phi(\vx)\), the chain rule gives
\begin{equation}
\nabla(f\circ\phi^{-1})(\vz;\xi)-\nabla(F\circ\phi^{-1})(\vz)
=J(\vz)^\top\delta_x(\vx;\xi).
\end{equation}
Therefore, with \(\Sigma_x(\vx)\defeq\E_{\xi}[\delta_x(\vx;\xi)\delta_x(\vx;\xi)^\top]\),
\begin{equation}
\E_{\xi}\!\left[\delta_x(\vx;\xi)^\top J(\vz)J(\vz)^\top\delta_x(\vx;\xi)\right]
=\operatorname{tr}\!\left(J(\vz)J(\vz)^\top\Sigma_x(\vx)\right),
\label{eq:app-sigma-coordinate-relation}
\end{equation}
while the pointwise \(x\)-coordinate variance is \(\operatorname{tr}(\Sigma_x(\vx))\). If \(m_\phi\mI\preceq J(\vz)J(\vz)^\top\preceq B_\phi^2\mI\), then
\begin{equation}
m_\phi\operatorname{tr}(\Sigma_x(\vx))\le\E_{\xi}\!\left[\delta_x(\vx;\xi)^\top J(\vz)J(\vz)^\top\delta_x(\vx;\xi)\right]\le B_\phi^2\operatorname{tr}(\Sigma_x(\vx)).
\end{equation}
Taking suprema over paired domains shows that the two noise floors are comparable only up to the local expander conditioning and the anisotropy of \(\Sigma_x\).

The shared-sample Rademacher bound used by both coordinates follows from the same moment calculation. Let \(G(\vy)=\E_\xi\!\left[G(\vy;\xi)\right]\), \(\vp\defeq\nabla G(\vy)\), \(\vp(\xi)\defeq\nabla G(\vy;\xi)\), and \(\sigma_G^2(\vy)\defeq\E_\xi[\|\vp(\xi)-\vp\|^2]\). Applying Eq.~\eqref{eq:smooth-rademacher-remainder} to \(G(\cdot;\xi)\) with sample-path smoothness \(L_G\) gives
\begin{equation}
\hgrad(G)(\vy;\vr)=\langle \vp(\xi),\vr\rangle\vr+\tau(\xi,\vr),
\qquad
\|\tau(\xi,\vr)\|\lesssim L_G\delta d^{3/2}.
\end{equation}
Let \(A_K\defeq K^{-1}\sum_{k=1}^K\vr_k\vr_k^\top\). Since
\begin{equation}
\E_{\vr}\!\left[A_K\right]=\mI,
\qquad
\E_{\vr}\!\left[A_K^2\right]=\left(1+\frac{d-1}{K}\right)\mI,
\end{equation}
and \(\E_\xi[\vp(\xi)]=\vp\), averaging \(K\) independent directions with one shared \(\xi\) yields the exact identity
\begin{equation}
\E_{\xi,\vr}\!\left[
\left\|
\frac1K\sum_{k=1}^K\langle\vp(\xi),\vr_k\rangle\vr_k-\vp
\right\|^2
\right]
=
\frac{d-1}{K}\|\vp\|^2
+\left(1+\frac{d-1}{K}\right)\sigma_G^2(\vy).
\label{eq:shared-sample-rademacher-decomposition}
\end{equation}
The first term is ordinary direction sampling. The second is the fluctuation of the shared sample path, and its leading component remains as \(K\to\infty\) because all directions use the same \(\xi\). Adding the Taylor remainder gives
\begin{equation}
\E\!\left[
\left\|
\frac1K\sum_{k=1}^K\hgrad(G)(\vy;\vr_k)-\nabla G(\vy)
\right\|^2
\right]
\lesssim
\frac{d}{K}\|\nabla G(\vy)\|^2
+L_G^2\delta^2d^3
+\left(1+\frac{d}{K}\right)\sigma_G^2(\vy).
\label{eq:shared-sample-rademacher-bound}
\end{equation}

For weight-space quantized querying, add and subtract the unquantized sampled finite-difference estimator:
\begin{equation}
\begin{aligned}
\hgrad(F\circ Q)(\vx)-\nabla F(\vx)
&=
\left(
\frac1K\sum_{k=1}^K
\frac{f(\vx+\mu\vu_k;\xi)-f(\vx-\mu\vu_k;\xi)}{2\mu}\vu_k
-\nabla F(\vx)
\right)\\
&\quad+\frac{1}{2\mu K}\sum_{k=1}^K
\left[f(Q(\vx+\mu\vu_k);\xi)-f(\vx+\mu\vu_k;\xi)\right]\vu_k\\
&\quad-\frac{1}{2\mu K}\sum_{k=1}^K
\left[f(Q(\vx-\mu\vu_k);\xi)-f(\vx-\mu\vu_k;\xi)\right]\vu_k .
\end{aligned}
\end{equation}
The first parenthesized term is the unquantized sampled estimator using the same shared \(\xi\). It is controlled by Eq.~\eqref{eq:shared-sample-rademacher-bound} with \(G(\cdot;\xi)=f(\cdot;\xi)\), \(L_G=L_x\), \(\delta=\mu\), and the domain-level bound \(\sigma_G^2\le\sigma_x^2\). It remains to bound the sampled endpoint-rounding residual. Assumption~\ref{assump:interior-grid} gives
\begin{equation}
\|Q(\vy)-\vy\|\le \frac{B_\phi\sqrt d\,\Delta}{2}
\end{equation}
for every probed point \(\vy\in\mathcal X\). Any point on the segment between \(\vx+s\mu\vu\) and \(Q(\vx+s\mu\vu)\) is within distance \(\mu\sqrt d+B_\phi\sqrt d\,\Delta/2\) of \(\vx\). Therefore, using the integral form of the fundamental theorem of calculus, Jensen's inequality, \(L_x\)-smoothness of \(F\), and the domain-level variance bound, for \(s\in\{-1,+1\}\),
\begin{equation}
\E_{\xi}\!\left[\left|f(Q(\vx+s\mu\vu);\xi)-f(\vx+s\mu\vu;\xi)\right|^2\right]\lesssim B_\phi^2d\Delta^2\left(\|\nabla F(\vx)\|^2+L_x^2d(\mu^2+B_\phi^2\Delta^2)+\sigma_x^2\right).
\label{eq:sampled-endpoint-pointwise-bound}
\end{equation}
Since \(\|\vu\|^2=d\), Jensen's inequality over the \(K\) directions gives
\begin{equation}
\begin{aligned}
&\E\!\left[
\left\|
\frac1K\sum_{k=1}^K
\frac{
f(Q(\vx+\mu\vu_k);\xi)-f(\vx+\mu\vu_k;\xi)
-f(Q(\vx-\mu\vu_k);\xi)+f(\vx-\mu\vu_k;\xi)
}{2\mu}\vu_k
\right\|^2
\right]\\
&\quad\lesssim \frac{B_\phi^2d^2\Delta^2}{\mu^2}\left(\|\nabla F(\vx)\|^2+L_x^2d(\mu^2+B_\phi^2\Delta^2)+\sigma_x^2\right).
\end{aligned}
\label{eq:sampled-endpoint-simple-bound}
\end{equation}
Combining Eq.~\eqref{eq:shared-sample-rademacher-bound} with Eq.~\eqref{eq:sampled-endpoint-simple-bound} gives
\begin{equation}
\begin{aligned}
\E\!\left[\left\|
\hgrad(F\circ Q)(\vx)-\nabla F(\vx)
\right\|^2\right]
&\lesssim
\frac{d}{K}\|\nabla F(\vx)\|^2+L_x^2\mu^2d^3+\left(1+\frac{d}{K}\right)\sigma_x^2\\
&\quad+\frac{B_\phi^2d^2\Delta^2}{\mu^2}\left(\|\nabla F(\vx)\|^2+L_x^2d(\mu^2+B_\phi^2\Delta^2)+\sigma_x^2\right).
\end{aligned}
\label{eq:weight-sampled-loss-bound}
\end{equation}
For a scalar compander coordinate, \(\rho\approx\mu|\nabla\phi(y)u|/\Delta\) for small \(\mu\). Thus the \(\Delta^2/\mu^2\) term in Eq.~\eqref{eq:weight-sampled-loss-bound} is the simplified \(1/\rho^2\)-type under-resolution penalty diagnosed by Section~\ref{sec:quantized-zo}.

For CAQ-ZO, the on-grid identity removes the sampled coupling term before the stochastic analysis begins. Applying Eq.~\eqref{eq:shared-sample-rademacher-bound} with \(G(\vz;\xi)=f(\phi^{-1}(\vz);\xi)\), \(L_G=L_z\), \(\delta=\Delta\), and the domain-level bound \(\sigma_G^2\le\sigma_z^2\) gives
\begin{equation}
\begin{aligned}
\E\!\left[\left\|
\hgrad(F\circ\phi^{-1})(\vz)-\nabla(F\circ\phi^{-1})(\vz)
\right\|^2\right]
&\lesssim
\frac{d}{K}\|\nabla(F\circ\phi^{-1})(\vz)\|^2
+L_z^2\Delta^2d^3\\
&\quad+\left(1+\frac{d}{K}\right)\sigma_z^2.
\end{aligned}
\label{eq:caq-sampled-loss-direct-bound}
\end{equation}
Equations~\eqref{eq:weight-sampled-loss-bound} and~\eqref{eq:caq-sampled-loss-direct-bound} prove Eq.~\eqref{eq:weight-estimator-budget} and Eq.~\eqref{eq:caq-estimator-budget}. If the oracle is deterministic, \(\sigma_x=\sigma_z=0\) and the sampled terms vanish; the remaining distinction is the simple \(\Delta^2/\mu^2\) endpoint-rounding term for weight-space queries versus zero for CAQ-ZO.

\subsubsection{Proof of Theorem~\ref{thm:caq-stationarity}}
\label{app:caq-stationarity-proof}

The two stationarity statements are proved separately. For the weight-space baseline, let
\begin{equation}
\vp_t\defeq\nabla F(\vx_t),\qquad
\vg_t\defeq\hgrad(F\circ Q)(\vx_t).
\end{equation}
The direct update is \(\vx_{t+1}=\vx_t-\eta\vg_t\). By \(L_x\)-smoothness of \(F\),
\begin{equation}
F(\vx_{t+1})
\le
F(\vx_t)
-\eta\langle\vp_t,\vg_t\rangle
+\frac{L_x\eta^2}{2}\|\vg_t\|^2 .
\label{eq:weight-space-descent-start}
\end{equation}
The conditional bias \(\vb_t\defeq\E\!\left[\vg_t\mid\vx_t\right]-\vp_t\) contains the ordinary centered finite-difference bias and the population endpoint-rounding residual. Combining Eq.~\eqref{eq:smooth-rademacher-remainder} with Eq.~\eqref{eq:weight-response-budget} gives
\begin{equation}
\|\vb_t\|^2
\lesssim
L_x^2\mu^2d^3
+\frac{B_\phi^2d^2\Delta^2}{\mu^2}
\Bigl[\|\vp_t\|^2+L_x^2d(\mu^2+B_\phi^2\Delta^2)\Bigr].
\label{eq:weight-space-bias-bound}
\end{equation}
Thus, if \(B_\phi^2d^2\Delta^2/\mu^2\le c_{\text{\normalfont res}}\) for a sufficiently small universal constant, Young's inequality gives
\begin{equation}
-\eta\left\langle\vp_t,\E\!\left[\vg_t\mid\vx_t\right]\right\rangle
\le
-\frac{3\eta}{4}\|\vp_t\|^2
+C\eta\left(L_x^2\mu^2d^3+\frac{B_\phi^2d^2\Delta^2}{\mu^2}L_x^2d(\mu^2+B_\phi^2\Delta^2)\right).
\label{eq:weight-space-biased-inner-product}
\end{equation}
The same estimator-error bound also controls the second moment:
\begin{equation}
\begin{aligned}
\E\!\left[\|\vg_t\|^2\mid\vx_t\right]
&\lesssim
\left(1+\frac{d}{K}+\frac{B_\phi^2d^2\Delta^2}{\mu^2}\right)\|\vp_t\|^2\\
&\quad+
L_x^2\mu^2d^3+\left(1+\frac{d}{K}\right)\sigma_x^2
+\frac{B_\phi^2d^2\Delta^2}{\mu^2}
\Bigl[L_x^2d(\mu^2+B_\phi^2\Delta^2)+\sigma_x^2\Bigr].
\end{aligned}
\label{eq:weight-space-second-moment}
\end{equation}
In the descent inequality, the deterministic part of \(L_x\eta^2\E[\|\vg_t\|^2\mid\vx_t]\) also produces
\[
L_x\eta^2L_x^2\mu^2d^3
\quad\text{and}\quad
L_x\eta^2\frac{B_\phi^2d^2\Delta^2}{\mu^2}L_x^2d(\mu^2+B_\phi^2\Delta^2).
\]
The step-size condition below implies \(L_x\eta\lesssim1\), so these terms are bounded by the corresponding \(C\eta\)-scaled bias-floor terms in Eq.~\eqref{eq:weight-space-biased-inner-product}. The current-gradient part of Eq.~\eqref{eq:weight-space-second-moment} is absorbed into the descent decrease by the same step-size restriction.
Taking conditional expectation in Eq.~\eqref{eq:weight-space-descent-start} and applying Eqs.~\eqref{eq:weight-space-biased-inner-product}--\eqref{eq:weight-space-second-moment} gives
\begin{equation}
\begin{aligned}
\E\!\left[F(\vx_{t+1})\mid\vx_t\right]
&\le
F(\vx_t)-c_0\eta\|\vp_t\|^2\\
&\quad+C\eta\left[
L_x^2\mu^2d^3
+\frac{B_\phi^2d^2\Delta^2}{\mu^2}L_x^2d(\mu^2+B_\phi^2\Delta^2)
\right]\\
&\quad+C L_x\eta^2\left(1+\frac{d}{K}+\frac{B_\phi^2d^2\Delta^2}{\mu^2}\right)\sigma_x^2,
\end{aligned}
\end{equation}
provided \(\eta\le c_\eta/[L_x(1+d/K+B_\phi^2d^2\Delta^2/\mu^2)]\). Summing this inequality, taking full expectation, and using the lower bound \(F_\star\) proves Eq.~\eqref{eq:weight-space-stationarity-budget}.

For the CAQ-ZO part, let
\begin{equation}
H(\vz)\defeq F(\phi^{-1}(\vz)),\qquad
\vp_t\defeq\nabla H(\vz_t),\qquad
\vg_t\defeq\hgrad_t(F\circ\phi^{-1}).
\end{equation}
The CAQ-ZO update can be written as
\begin{equation}
\bar{\vz}_{t+1}\defeq \vz_t-\eta\vg_t,\qquad
\vz_{t+1}=U(\bar{\vz}_{t+1})=\bar{\vz}_{t+1}+\mathbf q_t,
\end{equation}
where round-to-nearest on the \(z\)-grid gives
\begin{equation}
\|\mathbf q_t\|^2\le \frac{d\Delta^2}{4}.
\label{eq:update-projection-norm}
\end{equation}
This bound is the point in the proof that requires the pre-projection no-clipping condition in Theorem~\ref{thm:caq-stationarity}: \(\bar{\vz}_{t+1}\) lies in the unclipped round-to-nearest region, and the segment from \(\bar{\vz}_{t+1}\) to \(U(\bar{\vz}_{t+1})\) remains in \(\mathcal Z\).
By \(L_z\)-smoothness of \(H\),
\begin{equation}
H(\vz_{t+1})
\le
H(\vz_t)
{}+\langle \vp_t,-\eta\vg_t+\mathbf q_t\rangle
{}+\frac{L_z}{2}\|-\eta\vg_t+\mathbf q_t\|^2 .
\end{equation}
Using \(\|\va+\vb\|^2\le2\|\va\|^2+2\|\vb\|^2\) and Young's inequality on \(\langle \vp_t,\mathbf q_t\rangle\),
\begin{equation}
\begin{aligned}
H(\vz_{t+1})
&\le
H(\vz_t)
-\eta\langle\vp_t,\vg_t\rangle
{}+\frac{\eta}{8}\|\vp_t\|^2
{}+L_z\eta^2\|\vg_t\|^2
{}+\left(\frac{2}{\eta}+L_z\right)\|\mathbf q_t\|^2 .
\end{aligned}
\label{eq:descent-with-projection}
\end{equation}
It remains to control the two estimator terms. The sampled-loss fluctuation is centered by the definition \(F(\vx)=\E_{\xi}\!\left[f(\vx;\xi)\right]\), so the smoothness calculation in Eq.~\eqref{eq:smooth-rademacher-remainder} implies that the conditional bias
\begin{equation}
\vb_t\defeq \E\!\left[\vg_t\mid \vz_t\right]-\vp_t
\end{equation}
satisfies
\begin{equation}
\|\vb_t\|^2\le C L_z^2\Delta^2d^3.
\label{eq:caq-bias-bound}
\end{equation}
Therefore
\begin{equation}
-\eta\left\langle\vp_t,\E\!\left[\vg_t\mid\vz_t\right]\right\rangle
=-\eta\|\vp_t\|^2-\eta\langle\vp_t,\vb_t\rangle
\le -\frac{7\eta}{8}\|\vp_t\|^2+C\eta L_z^2\Delta^2d^3.
\label{eq:biased-inner-product}
\end{equation}
The estimator error bound in Eq.~\eqref{eq:caq-estimator-budget} also gives
\begin{equation}
\E\!\left[\|\vg_t\|^2\mid\vz_t\right]
\le
C\left(1+\frac{d}{K}\right)\|\vp_t\|^2
+C L_z^2\Delta^2d^3
+C\left(1+\frac{d}{K}\right)\sigma_z^2.
\label{eq:caq-second-moment}
\end{equation}
The first term in Eq.~\eqref{eq:caq-second-moment} is proportional to the current population gradient norm and is not replaced by a uniform sample-gradient bound. It is the multiplicative second-moment term from random directions; the step-size restriction below absorbs it into the descent decrease.
The deterministic finite-difference part \(L_z\eta^2L_z^2\Delta^2d^3\) is also folded into the \(C\eta L_z^2\Delta^2d^3\) bias floor because the same restriction gives \(L_z\eta\lesssim1\).
Taking conditional expectation in Eq.~\eqref{eq:descent-with-projection}, then applying Eqs.~\eqref{eq:update-projection-norm}--\eqref{eq:caq-second-moment}, yields
\begin{equation}
\E\!\left[H(\vz_{t+1})\mid\vz_t\right]
\le
H(\vz_t)
-c_0\eta\|\vp_t\|^2
+C\eta L_z^2\Delta^2d^3
+C L_z\eta^2\left(1+\frac{d}{K}\right)\sigma_z^2
+C\frac{d\Delta^2}{\eta},
\end{equation}
provided \(\eta\le c_\eta/[L_z(1+d/K)]\), with \(c_\eta\) a sufficiently small universal constant. The term \(L_zd\Delta^2\) is absorbed into \(Cd\Delta^2/\eta\) under \(\eta L_z\le 1\).

Taking full expectation, summing from \(t=0\) to \(T-1\), and using the lower bound \(H_\star=(F\circ\phi^{-1})_\star\) gives
\begin{equation}
c_0\eta\sum_{t=0}^{T-1}\E\!\left[\|\nabla H(\vz_t)\|^2\right]
\le
H(\vz_0)-H_\star
+CT\eta L_z^2\Delta^2d^3
+CT L_z\eta^2\left(1+\frac{d}{K}\right)\sigma_z^2
+CT\frac{d\Delta^2}{\eta}.
\end{equation}
Dividing by \(c_0\eta T\) and absorbing \(1/c_0\) into \(C\) proves Eq.~\eqref{eq:caq-stationarity-budget}.

\subsubsection{Proof of Theorem~\ref{thm:caq-weight-space-stationarity}}
\label{app:caq-weight-space-transfer-proof}

Let \(J(\vz)\defeq\nabla\phi^{-1}(\vz)\). The chain rule gives
\begin{equation}
\nabla(F\circ\phi^{-1})(\vz)=J(\vz)^\top\nabla F(\phi^{-1}(\vz)).
\end{equation}
With \(\vx_t=\phi^{-1}(\vz_t)\),
\begin{equation}
\|\nabla(F\circ\phi^{-1})(\vz_t)\|^2
=
\nabla F(\vx_t)^\top J(\vz_t)J(\vz_t)^\top\nabla F(\vx_t).
\end{equation}
The assumed local nonsingularity \(J(\vz_t)J(\vz_t)^\top\succeq m_\phi\mI\) implies
\begin{equation}
\|\nabla F(\vx_t)\|^2
\le
\frac{1}{m_\phi}\|\nabla(F\circ\phi^{-1})(\vz_t)\|^2 .
\end{equation}
Averaging this inequality over \(t\), taking expectation, and substituting Eq.~\eqref{eq:caq-stationarity-budget} proves Eq.~\eqref{eq:caq-weight-space-transfer}. The factor \(C/m_\phi\) is therefore the universal constant from Theorem~\ref{thm:caq-stationarity} multiplied by the expander conditioning penalty \(1/m_\phi\).

\section{Auxiliary Query-Geometry Calculations}
\label{app:auxiliary-bounds}

Appendix~\ref{app:proofs} gives the theorem proofs in dependency order. The following calculations make three auxiliary effects explicit: a standalone weight-space coupling bound, the off-grid compander-coordinate contrast, and the descent calculation showing why endpoint rounding enters stationarity as a bias condition rather than only as a variance term.

\subsection{Standalone Weight-Space Coupling Bound}
\label{app:weight-space-coupling}

Let
\begin{equation}
\ve_\pm\defeq Q(\vx\pm\mu\vu)-(\vx\pm\mu\vu)
\end{equation}
be the induced weight-domain quantization residual at the two perturbed query points. The quantization-specific difference between the quantized weight-space response and the ordinary weight-space response is
\begin{equation}
\begin{aligned}
\hgrad(F\circ Q)(\vx;\vu)-\hgrad(F)(\vx;\vu)
&=
\frac{F(\vx+\mu\vu+\ve_+)-F(\vx+\mu\vu)}{2\mu}\vu\\
&\quad-
\frac{F(\vx-\mu\vu+\ve_-)-F(\vx-\mu\vu)}{2\mu}\vu .
\end{aligned}
\end{equation}
Under the same \(L_x\)-smoothness convention used in the main theorem,
\begin{equation}
|F(\vy+\ve)-F(\vy)|^2\lesssim \|\ve\|^2\left(\|\nabla F(\vx)\|^2+L_x^2\|\vy-\vx\|^2+L_x^2\|\ve\|^2\right)
\end{equation}
for \(\vy=\vx\pm\mu\vu\) and \(\ve=\ve_\pm\), with the gradient on the segment between \(\vy\) and \(\vy+\ve\). Combining this inequality with \(\|\vu\|=\sqrt d\), \(\|\vy-\vx\|=\mu\sqrt d\), and \(\|\ve_\pm\|\le B_\phi\sqrt d\,\Delta/2\) gives
\begin{equation}
\|\hgrad(F\circ Q)(\vx;\vu)-\hgrad(F)(\vx;\vu)\|^2\lesssim \frac{B_\phi^2d^2\Delta^2}{\mu^2}\left(\|\nabla F(\vx)\|^2+L_x^2d(\mu^2+B_\phi^2\Delta^2)\right).
\end{equation}
This is the smoothness-based form of the estimator-level identity in Eq.~\eqref{eq:coupling-term}. Under a scalar compander away from clipping, the mean-value theorem gives \(|Q(y)-y|\le\Delta/(2\nabla\phi(\bar y))\) for some \(\bar y\) between \(y\) and \(Q(y)\), connecting the bound to the local cell length described in Section~\ref{sec:quantized-zo}.

\subsection{Off-Grid Compander-Coordinate Querying}
\label{app:compressor-reduction}

CAQ-ZO uses \(z\)-coordinate endpoints that already lie on the grid. If a compander-coordinate method instead queries off-grid points \(\vz\pm\mu\vu\), the same endpoint-rounding channel reappears in the \(z\)-coordinate. Write \(H\defeq F\circ\phi^{-1}\) and \(\ve_\pm\defeq U(\vz\pm\mu\vu)-(\vz\pm\mu\vu)\). The residual bound \(\|\ve_\pm\|_\infty\le\Delta/2\) implies \(\|\ve_\pm\|\le \sqrt d\,\Delta/2\), and the response gap is
\begin{equation}
\begin{aligned}
\hgrad(H\circ U)(\vz;\vu)-\hgrad H(\vz;\vu)
=\frac{1}{2\mu}\Big(&
H(\vz+\mu\vu+\ve_+)-H(\vz+\mu\vu)\\
&-H(\vz-\mu\vu+\ve_-)+H(\vz-\mu\vu)
\Big)\vu .
\end{aligned}
\end{equation}
On a local region where \(H\) is \(L\)-smooth, the same smoothness-integral argument as Eq.~\eqref{eq:offgrid-response-budget} gives
\begin{equation}
\E\!\left[\|\hgrad(H\circ U)(\vz;\vu)-\hgrad H(\vz;\vu)\|^2\right]
\le C\frac{d^2\Delta^2}{\mu^2}\left(\|\nabla H(\vz)\|^2+L^2d(\mu^2+\Delta^2)\right),
\end{equation}
where \(C\) is a universal numerical constant and the displayed dimension factors come from the residual norm, the finite-difference direction norm, and the local smoothness radius. This contrast isolates the point that changing coordinates is insufficient unless the endpoints are also aligned with the grid. If residuals are centered by dithering or stochastic rounding, part of this term can be averaged over directions, but the theorem statements are for deterministic round-to-nearest quantization and do not assume that additional centering.

\subsection{Descent Calculation for Inexact ZO}
\label{app:transformed-convergence}

The stationarity certificate in Theorem~\ref{thm:caq-stationarity} uses a smooth nonconvex descent calculation with one distinction between second moment and bias. Let \(G\) be the smooth objective in the coordinate being updated, let \(\hgrad_t(G)\) be the measured finite-difference estimator at \(\vy_t\), and let \(\vy_{t+1/2}=\vy_t-\eta\hgrad_t(G)\) be the pre-projection update. \(L\)-smoothness gives
\begin{equation}
G(\vy_{t+1/2})
\le
G(\vy_t)
-\eta\langle\nabla G(\vy_t),\hgrad_t(G)\rangle
+\frac{L\eta^2}{2}\|\hgrad_t(G)\|^2 .
\end{equation}
The \(L\eta^2\|\hgrad_t(G)\|^2\) term can absorb multiplicative current-gradient contributions in the estimator second moment by reducing the step size. The inner-product term is different: if the estimator has conditional bias
\(\vb_t\defeq\E\!\left[\hgrad_t(G)\mid\vy_t\right]-\nabla G(\vy_t)\), the descent argument requires this bias to be aligned with descent or small relative to \(\|\nabla G(\vy_t)\|\). Let \(B_{\text{\normalfont bias}}\) collect the non-gradient terms from this conditional-bias control and let \(B_{\text{\normalfont var}}\) collect sampled-loss variance terms that enter only through the second moment. Young's inequality and total expectation yield
\begin{equation}
\E\!\left[G(\vy_{t+1/2})\right]
\le
\E\!\left[G(\vy_t)\right]
-c\eta\,\E\!\left[\|\nabla G(\vy_t)\|^2\right]
+C\eta B_{\text{\normalfont bias}}+CL\eta^2B_{\text{\normalfont var}}
\end{equation}
for universal constants \(c,C>0\), with the step-size restriction absorbing only the second-moment multiplier. Summing over \(t\) gives
\begin{equation}
\frac1T\sum_{t=0}^{T-1}\E\!\left[\|\nabla G(\vy_t)\|^2\right]
\le
 C\frac{G(\vy_0)-G_\star}{\eta T}
+CB_{\text{\normalfont bias}}+CL\eta B_{\text{\normalfont var}}+C\frac{d\Delta^2}{\eta^2}
\end{equation}
for nearest-grid projection with spacing \(\Delta\). If there is no storage projection, the last term is omitted. For CAQ-ZO, \(G=F\circ\phi^{-1}\), \(B_{\text{\normalfont bias}}=L_z^2\Delta^2d^3\), and \(B_{\text{\normalfont var}}=(1+d/K)\sigma_z^2\). For weight-space querying, endpoint rounding enters the conditional bias; this is why Theorem~\ref{thm:caq-stationarity} imposes \(B_\phi^2d^2\Delta^2/\mu^2\le c_{\text{\normalfont res}}\) rather than treating the current-gradient residual as a step-size-only term.

\section{Compander Families and Modeling Scope}
\label{app:compander-families}

The compander assumption covers scalar nonuniform quantizers that can be represented, exactly or approximately, as \(Q=\phi^{-1}\circ U\circ\phi\). It is not a claim that every nonuniform codebook has a canonical smooth scalar \(z\)-coordinate.

\paragraph{Exact companders.}
Uniform INT quantization is the identity special case \(\phi=\mathrm{id}\). Classical logarithmic companders are exact members of the model class. For example, with strength \(c>0\) and \(|x|\le\alpha\),
\begin{equation}
\phi_c(x)=
\sgn(x)\frac{\log(1+c |x|/\alpha)}{\log(1+c)},
\qquad
\phi_c^{-1}(z)=
\alpha\,\sgn(z)\frac{(1+c)^{|z|}-1}{c}.
\end{equation}
A-law companding is also exact, with a piecewise-linear/log map:
\begin{equation}
\phi_A(x)=\sgn(x)
\begin{cases}
\dfrac{A|x|/\alpha}{1+\log A}, & |x|/\alpha<1/A,\\[4pt]
\dfrac{1+\log(A|x|/\alpha)}{1+\log A}, & 1/A\le |x|/\alpha\le 1 .
\end{cases}
\end{equation}
Distribution-aware scalar quantizers also fit the framework when \(\phi(x)\) is a smooth monotone CDF or spline, for example \(\phi(x)=F_X(x)\) and \(\phi^{-1}(z)=F_X^{-1}(z)\).

\paragraph{Table-codebook abstractions.}
For LLM fine-tuning, NF4 is the most relevant nonuniform weight format in the experiments. The exact theorem statements apply to exact smooth companders. A table-based codebook such as bitsandbytes NF4 can be connected to the theory by fitting a monotone interpolation to the codebook, for example through a Gaussian quantile abstraction with scale \(s>0\),
\begin{equation}
\phi_{\text{\normalfont NF4}}(x)\approx \Phi(x/s),
\qquad
\phi_{\text{\normalfont NF4}}^{-1}(z)\approx s\,\Phi^{-1}(z).
\end{equation}
This interpolation introduces an additional approximation term measuring the mismatch between the implemented table and the smooth compander. NF4-style formats are therefore motivating abstractions rather than exact members of the theorem class.

\subsection{Beyond Scalar Companders}
\label{app:beyond-companders}

APoT-style additive powers-of-two codebooks, arbitrary learned k-means codebooks, and vector quantizers do not generally admit a single smooth scalar \(\phi\) whose inverse makes the codebook uniform. A local relaxation can use neighboring codebook gaps as local perturbation radii, which may reduce coupling in nearly uniform regions but does not eliminate it globally. A more direct codebook-aware estimator would sample perturbation directions from neighbor differences \(v\in\{c_j-c_i:c_j\in\mathcal N(c_i)\}\), guaranteeing queried points are codebook points at the cost of asymmetric and finite direction sets.

Vector quantizers require a different query geometry. If each block is quantized to a codebook \(\mathcal C\subset\R^m\), on-grid endpoint alignment becomes a high-dimensional codebook-neighbor problem, not a scalar compander problem. Lattice-structured vector quantizers may admit an analogue of on-grid alignment, but arbitrary vector codebooks do not. In mixed-precision models, a practical compromise is layerwise: use CAQ-ZO where a scalar compander is available, the same rule with \(\phi=\mathrm{id}\) for uniform low-bit layers, and weight-space ZO where neither scalar grid structure nor a reliable codebook-neighbor geometry is available.

\section{On-Grid Estimator Details}
\label{app:on-grid-details}

On-grid querying enters the formal results at two points: it makes endpoint quantization the identity in Theorem~\ref{thm:endpoint-rounding-response}, and it lets the CAQ-ZO estimator reduce to an ordinary Rademacher finite difference for \(F\circ\phi^{-1}\) in Theorem~\ref{thm:estimator-error-query-geometry}. This section isolates that structural identity and the two small calculations needed to interpret the special uniform-grid case and the \(z\)-to-weight stationarity transfer.

\begin{lemma}[On-grid query identity]
\label{lem:on-grid-identity}
For uniform quantization with grid \(\mathcal G\) and spacing \(\Delta\), if \(\vx\in\mathcal G\), \(\vr\in\{-1,+1\}^d\), and no clipping is activated, then \(\vx\pm\Delta\vr\in\mathcal G\) and \(Q(\vx\pm\Delta\vr)=\vx\pm\Delta\vr\). For a companding quantizer, the same statement holds in the \(z\)-coordinate: if \(\vz\in\mathcal G\), then \(Q(\phi^{-1}(\vz\pm\Delta\vr))=\phi^{-1}(\vz\pm\Delta\vr)\).
\end{lemma}

This lemma is the structural basis of on-grid querying. It removes the quantizer from the function evaluation rather than only bounding its residual.

\subsection{Uniform-Grid Special Case}
\label{app:uniform-grid-special-case}

When \(\phi=\mathrm{id}\), CAQ-ZO reduces to the uniform-grid aligned Rademacher estimator. For a grid point \(\vx\), define the two-point estimator
\begin{equation}
\hgrad(F)(\vx;\vr)
\defeq
\frac{F(\vx+\Delta\vr)-F(\vx-\Delta\vr)}{2\Delta}\,\vr,
\end{equation}
the \(L_x\)-smooth finite-difference bound gives
\begin{equation}
\left\|\E\!\left[\hgrad(F)(\vx;\vr)\right]-\nabla F(\vx)\right\|^2
\le C\Delta^2L_x^2d^3 .
\end{equation}
Combining Rademacher orthogonality with the bias bound gives, for \(\hgrad(F)(\vx)\) defined as the average of \(K\) independent copies,
\begin{equation}
\E\!\left[\|\hgrad(F)(\vx)-\nabla F(\vx)\|^2\right]
\le C
\frac{d\|\nabla F(\vx)\|^2}{K}
{}+C\Delta^2L_x^2d^3 .
\end{equation}

\subsection{Coordinate-Conditioning Calculation}
\label{app:compressor-smoothness-cost}

For CAQ-ZO, smoothness is controlled by the composite \(H=F\circ\phi^{-1}\), not only by the derivatives of \(F\). The main convergence theorem assumes the local constant \(L_z\) directly for this composite objective, so it does not require a sampled-gradient bound. The following chain-rule calculation is only a sufficient way to interpret how expander curvature can affect this deterministic composite smoothness constant. Let \(J(\vz)=\nabla\phi^{-1}(\vz)\). The second-derivative chain rule gives
\begin{equation}
\nabla^2H(\vz)
=
J(\vz)^\top\nabla^2F(\vx)J(\vz)
+\sum_i \nabla_i F(\vx)\nabla^2\phi^{-1}_i(\vz),
\qquad \vx=\phi^{-1}(\vz).
\end{equation}
Taking operator norms under \(\|J\|\le B_1\), \(\|\nabla^2\phi^{-1}\|\le B_2\), and \(\|\nabla^2F\|\le L_F\) gives the local sufficient bound
\begin{equation}
L_z\le L_FB_1^2+B_2\sup_{\vx\in\mathcal X}\|\nabla F(\vx)\|.
\end{equation}

For the stationarity-transfer statement, the chain rule gives
\begin{equation}
\|\nabla H(\vz)\|^2
=
\nabla F(\vx)^\top\nabla\phi^{-1}(\vz)\nabla\phi^{-1}(\vz)^\top\nabla F(\vx),
\qquad \vx=\phi^{-1}(\vz).
\end{equation}
Applying \(m_\phi\mI\preceq\nabla\phi^{-1}(\vz)\nabla\phi^{-1}(\vz)^\top\preceq B_\phi^2\mI\) gives the gradient-norm transfer in Eq.~\eqref{eq:caq-weight-space-transfer}. Thus \(L_z\) can increase with expander curvature and conditioning when it is derived from separate bounds on \(F\) and \(\phi^{-1}\). The main theorem instead assumes the local composite smoothness constant \(L_z\) directly; this is the smooth-optimization price of making the query geometry exactly on-grid in the \(z\)-coordinate.

\section{Algorithmic Details}
\label{app:algorithm}

CAQ-ZO's compander-aligned query geometry uses two implementation invariants. First, the current iterate is represented by a \(z\)-coordinate grid point \(\vz_t\). Second, every evaluation point is generated as \(\vz_t\pm\Delta\vr\) before mapping back to \(\vx\). These two invariants make \(Q\) equal the identity at the queried points. After the transformed-coordinate update, the intermediate point \(\vz_{t+1/2}=\vz_t-\eta\hgrad_t(F\circ\phi^{-1})\) is quantized back to the \(z\)-grid. This update-time projection is distinct from query-time quantization-perturbation coupling: it can create a projection floor, but it does not invalidate the on-grid query identity.

For the uniform special case \(\phi=\mathrm{id}\), CAQ-ZO queries \(\vx_t\pm\Delta\vr\), where \(\Delta\) is the uniform grid spacing. This recovers the same on-grid identity without any coordinate transformation.

\section{Additional Empirical Details}
\label{app:experiments}

\subsection{Synthetic Estimator Residual}
\label{app:synthetic-details}

\begin{figure}[t]
\centering
\includegraphics[width=0.88\textwidth]{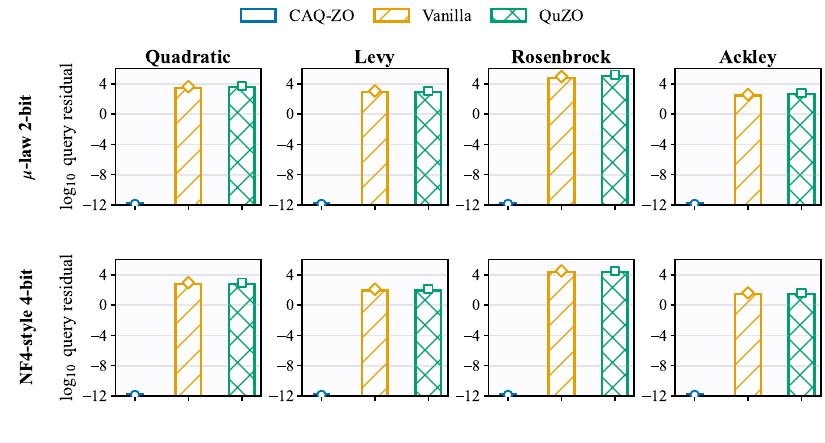}
\caption{Query-time estimator residual under the shared synthetic setting used by the main convergence figure. Rows show \(\mu\)-law 2-bit and NF4-style 4-bit companders; columns show Quadratic, Levy, Rosenbrock, and Ackley objectives. Each bar reports \(\log_{10}(\|\hat g_{\text{\normalfont meas}}-\hat g_{\text{\normalfont unrounded}}\|^2/\|g_\star\|^2)\) for one-step \(K=4\) estimators, averaged over three start-matched initializations and 32 independent probes per start, with mean \(\pm\,2\) standard errors. The target \(g_\star\) is \(\nabla(F\circ\phi^{-1})(\vz)\) for CAQ-ZO and \(\nabla F(\vx)\) for weight-space baselines. CAQ-ZO queries on-grid compander-coordinate endpoints, so its query-time residual is zero and plotted at the \(10^{-12}\) numerical floor.}
\label{fig:app-shared-residual}
\end{figure}

Figure~\ref{fig:app-shared-residual} directly measures the query-time residual that appears in Theorems~\ref{thm:endpoint-rounding-response} and~\ref{thm:estimator-error-query-geometry}. The setting is the same one used for the main synthetic convergence comparison: \(d=10000\), \(K=4\), \(T=10000\), three start-matched initializations, Quadratic/Levy/Rosenbrock/Ackley objectives, FP-master states, per-block scale calibration, dynamic scale recalibration, Adam, clipping, deterministic endpoint rounding, and fixed \(\eta=0.005\). Each one-step probe subtracts the unrounded estimator computed with identical directions, so the plotted quantity isolates endpoint rounding from optimizer state and random-direction sampling.

\subsection{LLM Fine-Tuning Details}
\label{app:llm-details}

The LLM fine-tuning experiment follows the few-shot sampling convention used in the MeZO evaluation suite: each task uses 1{,}000 training examples, 500 validation examples, and 1{,}000 test examples. Each table entry is the validation-selected test result for that method. SST-2, RTE, BoolQ, CB, and COPA report accuracy; SQuAD reports F1. Table~\ref{tab:hyperparam} lists the grid used for batch size and learning rate.

The Qwen-2.5-1.5B runs were executed on an NVIDIA RTX 4090 worker, and the Llama-2-7B runs were executed on an NVIDIA RTX A6000 worker. After selecting the hyperparameters from the validation grid, each selected LLM configuration was run at least twice. These repeats were used to check that the matched NF4 comparison was stable under the selected setting; the current table reports the validation-selected test results rather than confidence intervals or wall-clock measurements.

\begin{table}[t]
\centering
\caption{Hyperparameter search grids for the LLM fine-tuning experiment.}
\label{tab:hyperparam}
\small
\begin{tabular}{@{}llc@{}}
\toprule
Method & Hyperparameter & Values \\
\midrule
\multirow{2}{*}{MeZO}
 & Batch size    & $\{8,16\}$ \\
 & Learning rate & $\{10^{-5},\,10^{-6}\}$ \\
\midrule
\multirow{2}{*}{QuZO}
 & Batch size    & $\{8,16\}$ \\
 & Learning rate & $\{5\cdot10^{-5},\,10^{-5},\,10^{-6}\}$ \\
\midrule
\multirow{2}{*}{CAQ-ZO}
 & Batch size    & $\{8,16\}$ \\
 & Learning rate & $\{5\cdot10^{-5},\,10^{-5},\,10^{-6}\}$ \\
\bottomrule
\end{tabular}
\end{table}

The scoped Qwen/Llama NF4 fine-tuning table supports the real-task query-geometry comparison under matched quantizer and evaluation budget. Broader large-model claims would require seed aggregation, wall-clock measurements, memory measurements, and additional codebook families.

\section{Broader Impacts}
\label{app:impacts}

This work is foundational optimization research. Its main positive impact is to reduce the memory needed for model adaptation when gradients are unavailable or too expensive to store, which can make experimentation with large models more accessible. The same efficiency improvements could also lower the cost of adapting models for harmful or low-quality applications. The work does not release a model, dataset, or deployment system, and the controlled experiments use synthetic objectives; concrete downstream risks therefore depend on future applications of the optimizer.

\end{document}